\documentclass{article}

\usepackage{arxiv}

\usepackage[utf8]{inputenc} 
\usepackage[T1]{fontenc}    
\usepackage{hyperref}       
\usepackage{url}            
\usepackage{booktabs}       
\usepackage{amsfonts}       
\usepackage{nicefrac}       
\usepackage{microtype}      
\usepackage{lipsum}
\usepackage{graphicx}
\graphicspath{ {./images/} }
\usepackage{amsmath}
\usepackage{titlesec}
\usepackage{multirow}
\usepackage{appendix}
\usepackage{pifont}
\usepackage{amssymb}
\usepackage{etoolbox}
\usepackage{pifont}
\usepackage{graphicx}
\usepackage{subcaption}
\usepackage{enumitem}

\makeatletter
\renewenvironment{quote}
  {\list{}{\rightmargin=1pt\leftmargin=1pt}%
   \item\relax}
  {\endlist}
\makeatother

\titleformat{\paragraph}
  {\normalfont\large\bfseries}
  {\theparagraph}
  {0.5em}
  {}

\newcommand{\minititle}[1]{%
  \vspace{0.5em}%
  \noindent\textbf{#1}
}

\titlespacing*{\paragraph}{0pt}
  {1.0ex plus 0.2ex minus 0.2ex}
  {0.5em}

\title{How Affect Propagates among LLM Agents: Emergent Emotional Contagion in Crowd Simulation}

\author{
 Funda Durupinar \\
  Computer Science Department\\
  University of Massachusetts\\
  Boston, MA 02125 \\
  \texttt{funda.durupinarbabur@umb.edu} \\  
}

\begin{document}
\maketitle
\begin{abstract}
This paper studies the behavior of language models in a multi-agent crowd simulation, focusing on how affect propagates among agents that perceive and appraise one another. Each agent perceives its neighbors through visual, auditory, and tactile channels, then appraises these perceptions in light of its prompted personality profile, memory, current affective state, and situational context. Appraisal is carried out by an LLM, which updates the agent’s internal affective state and selects its outward expression. The architecture contains no hand-authored mechanism for directly transferring affective state between agents; instead, inter-agent influence arises through the perception–appraisal–expression loop. The agent representation draws on the Big Five personality model and Russell’s circumplex model of affect. To limit latency, low-level steering and navigation are handled by a conventional crowd simulator operating independently of the LLM-based cognitive layer.

We evaluate the architecture across five scenario environments spanning alarming, joyful, and neutral situations in different spatial layouts. The results show that the system produces emotional contagion dynamics with spatial, temporal, and personality-dependent structure in sparse, small crowds. Alarm spreads from seeded agents as a traveling front, the mean alarmed fraction settles at a nonzero plateau, and the distribution of prompted personality profiles determines whether an ambiguous alarm ignites panic and whether a provocation is interpreted as anger or fear. We further evaluate the appraisal step through controlled experiments across prompt variants, sampling temperatures, and four model backends, showing that the dynamics are backend-dependent.  
\end{abstract}


\section{Introduction}
 
Societies of LLM agents offer a potentially disruptive paradigm for computational social science~\cite{piao2025agentsociety}. By providing in silico environments that approximate human behavior~\cite{bail2024can},  they enable controlled experiments that would be logistically impossible or unethical with human subjects, and let researchers study complex, emergent social interactions. As a result, they have been adopted as human proxies in economics~\cite{horton2023large, jia2024can}, affective neuroscience~\cite{Yang2026}, social and political science~\cite{ashery2025emergent, yang2024llm},  emergency simulations~\cite{dang2025large, sultimov2026respond}, and social networks~\cite{gao2023s3}. 

Beyond modeling humans, these societies also serve as testbeds to study the behavioral tendencies, limitations, and emergent dynamics of AI agents.  LLM-driven agents have been used to populate virtual spaces~\cite{park2023generative, wang2023humanoid}; to study emergent behavior in domains such as autonomous driving~\cite{wang2024autonomous}, emergency response coordination~\cite{silva2025urgencyaware},  delivery and warehouse robotics~\cite{srivastava2025deliver}; and to trace how errors or unsafe actions cascade through multi-agent and multi-robot collaborations~\cite{huang2026propagating, xie2026spark}. 

Building on this second direction, this paper explores emotional contagion among LLM-driven agents in a dynamic crowd simulation. We equip each crowd agent with an LLM-based ``mind'' that governs cognitive appraisal, internal emotional state, and the selection of outward expressions. Agents perceive the observable behavior of nearby agents, appraise it, and update their own emotions and expressions accordingly. These changes then become visible to others, creating a reciprocal feedback loop that links the emotional dynamics of neighboring agents. We investigate whether these local interactions produce coherent spatial, temporal, and dispositional patterns of affect propagation, and whether they give rise to collective behaviors such as panic. 
 
To keep the agents' behavior grounded in human psychology, their affective states are built on established psychological models. Personality is represented using the Big Five model~\cite{goldberg2013alternative}, and emotion is encoded using Russell’s circumplex model of affect~\cite{russell1980circumplex}.  Our approach decouples high-level decision-making from low-level spatial control.  The LLM governs appraisal, emotion, and expression, while crowd dynamics such as geometric path planning, obstacle avoidance, and rigid-body physics are handled by an underlying simulator. Although this separation may restrict some forms of emergence, it avoids relying on zero-shot language models for real-time spatial and physical reasoning tasks that are beyond their current capabilities. Using this setup, we investigate whether and how affect propagates through the crowd. We formulate the following hypotheses:

\noindent \textbf{H1.} A neighbor's observable expressions predict a change in an agent's affective state, as social appraisal theory predicts for humans~\cite{manstead2001social, parkinson2011interpersonal}.
\begin{itemize}[leftmargin=*, itemsep=0pt]
\item[]\textbf{H1.a} Alarm cues lead to decreased valence and increased arousal.
\item [] \textbf{H1.b} Positive expressions lead to increased valence and arousal~\cite{barsade2002ripple}.
\item []\textbf{H1.c} In a calm setting with no salient cues, affect is not driven by contagion; the crowd appraises the benign setting directly.
\end{itemize}

\noindent\textbf{H2.} From a seed, alarm spreads as a wave-like front~\cite{rosenthal2015revealing, farkas2002mexican, helbing2000escape}. An agent's onset of alarm is predicted by its distance from the seed. 

\noindent\textbf{H3.} The crowd's alarmed fraction over time follows a compartmental epidemic model, such as a Susceptible-Infected-Susceptible (SIS) or Susceptible-Infected-Recovered (SIR)~\cite{hethcote2000mathematics} or the generalized threshold contagion that personality-based crowd models build on~\cite{dodds2005generalized}. This framing of emotion spread is standard in the crowd-simulation literature~\cite{Durupinar2016-va, basak2018using, vanhaeringen2023review}.

\noindent\textbf{H4.} Contagion is carried by the modalities through which agents perceive
others' expressions. Blocking perception of all modalities collapses transmission to a
floor, and modalities differ in how much contagion each carries.

\noindent\textbf{H5.} Crowd personality distribution determines emergent contagion. This hypothesis is based on the literature on crowd simulation~\cite{durupinar2009ocean, Durupinar2016-va}.

\begin{itemize}[leftmargin=*, itemsep=0pt]
\item [] \textbf{H5.a} As mean neuroticism rises, an ambiguous, unseeded stimulus
is likely to tip the crowd into self-amplifying panic. Such an effect is not observed in a stable crowd.
\item [] \textbf{H5.b} A minority of dispositionally calm agents reduces the panic the rest of the crowd catches.
\item[]\textbf{H5.c} Anger spreads only when a provocateur is present and the receiving
crowd is anger-prone, with low agreeableness. A baseline crowd appraises the identical
provocation as fear.
\end{itemize}

In addition to testing these hypotheses, we evaluate the robustness of the appraisal model across prompt formulations, sampling temperatures, and backend models. The results show that appraisal is largely stable across prompt and temperature variations, whereas its sensitivity to personality depends on the backend model and influences whether contagion emerges.


The findings of this study may support the development of psychologically varied crowds for games, film, and training simulations. The architecture may also inform systems beyond crowd simulation. As LLM agents are increasingly deployed in multi-agent environments, understanding how one agent's expressed state influences neighboring agents is important for assessing the stability of such systems. Additionally, pending validation against empirical human-crowd data, appraisal-driven crowd models may also contribute to evacuation and safety research by making the relationship between crowd composition and panic propagation explicit and testable.

\section{Related Work}

\subsection{Generative Agent Societies and Emergent Collective Behavior}
Starting with the pioneering work of Park et al.~\cite{park2023generative}, simulations of micro-societies using LLM-driven social agents as proxies for human behavior have substantially contributed to the relatively recent paradigm of computational social science~\cite{ziems2024can}. Extending beyond a small number of agents, such simulations have become ambitious enough to simulate agent civilizations comprising thousands of agents~\cite{al2024project}. Concordia~\cite{vezhnevets2023generative}
generalizes this line of work into a reusable library for generative
agent-based models, in which agents act by describing their intentions in
natural language and a Game Master translates these into effects within a
physically or digitally grounded environment.

Once such societies exist, the natural next question is what happens when many LLM agents interact, independent of any single agent's internal architecture. Talebirad et al.~\cite{talebirad2025wisdom} study whether aggregating independent LLM estimates reproduces the classic ``wisdom of crowds'' effect for vision-based estimation tasks. They show that the median aggregation of deterministic responses from a diverse set of models is the most effective approach. Wu et al.~\cite{wu2024shall} examine spontaneous cooperation among competing LLM agents, demonstrating that emergent social behavior can arise without explicit directions to cooperate. Piao et al.~\cite{piao2025agentsociety} scale this question to large simulated societies, showing that LLM-driven generative agents can reproduce recognizable patterns of human behavior at population scale.  Lin et al.~\cite{lin2025crowdllm} build digital populations of LLM agents augmented with generative models as a lower-cost substitute for recruiting human participants in crowdsourcing and marketing studies.


\subsection{Crowd Simulation with LLMs}
The aforementioned societies are different from crowd simulations as
they explore social dynamics but not the physical aspects of crowds. A
second line of work brings LLMs into crowd simulation. Since
crowd dynamics strain the current capabilities of zero-shot LLMs,
locomotion and navigation are handled by classic crowd simulation
techniques, data-driven approaches, or fine-tuned models. Liu et
al.~\cite{liu2025emergent} study emergent crowd dynamics in
language-driven multi-agent scenes, where an LLM periodically sets each
agent's goals and steering parameters informed by inter-agent dialogue,
while low-level locomotion remains force-based. Panayiotou et
al.~\cite{Panayiotou2026} use LLMs to bootstrap synthetic datasets for
crowd scenarios, encoding the spatial and temporal evolution of a crowd
as a time-expanded graph from which new scenarios are generated. Hwang
et al.~\cite{hwang2026crowdvla} tie perception and action together with
embodied vision-language-action agents for context-aware crowd
simulation, fine-tuning a vision-language model via Low-Rank Adaptation
(LoRA) for crowd navigation.

Gu et al.~\cite{gu2026research} similarly keep motion outside the LLM,
combining the OCEAN model with a dimensional emotion model and an LLM
cognitive core, and using an emotion-conditioned neural A* for
personality- and affect-dependent pathfinding. Their emotion dynamics
follow an authored update rule in which traits scale emotion updates,
as in~\cite{Durupinar2016-va}, while the LLM is confined to memory,
planning, and dialogue. Our work instead removes the authored rule and
studies how emotion propagates between agents through appraisal.
 
LLM-driven agents have also been deployed in safety-critical crowd
scenarios, where existing systems differ mainly in which layer the LLM
controls. Sultimov et
al.~\cite{sultimov2026respond} use LLM agents in disaster-response 
simulations to capture emergent evacuation and communication behavior.
Saravanan et al.~\cite{saravanan2025ai} combine semantic segmentation of
building blueprints with a social-force model, with the LLM supplying
situational narratives for fire evacuation planning. Dang et
al.~\cite{dang2025large} similarly drive a cellular-automata fire
evacuation model with LLM agents that carry personalized memory and
cognition, and Yang et al.~\cite{yang2025agents} add LLM reasoning to
agent-based evacuation models, reporting gains in behavioral realism over
classical rule-based models.

\subsection{Personality and Emotion in LLM Agents}
Because crowd- and society-level behavior is ultimately built from individual agents, a body of work targets how personality and emotion are represented in agents and how those representations play out when agents interact.  Jiang et al.~\cite{jiang2023evaluating} evaluate and induce personality traits in pre-trained language models. They introduce the Machine Personality Inventory (MPI), built from standardized Big Five items, and find that LLMs exhibit consistent, human-comparable trait profiles. Extending this from test responses to generated text, PersonaLLM~\cite{jiang2024personallm} assigns LLMs Big Five personas and shows that they not only report trait-consistent scores on the 44-item BFI but also produce stories whose linguistic patterns reflect the assigned traits, partly mirroring human writing.  Mao et al.~\cite{mao2024editing} instead treat personality as a model-editing target. Given a topic, they edit an LLM so its expressed opinion reflects a chosen trait, and introduce the PersonalityEdit benchmark over neuroticism, extraversion, and agreeableness. 

A parallel line asks whether LLMs process emotion in ways that resemble psychological theory. Tak and Gratch~\cite{tak2023gpt} probe whether an LLM's emotional judgments follow appraisal theory. They follow a component perspective on evaluation, which asserts that a model should be validated by evaluating each of its component processes. Broekens et al.~\cite{broekens2023fine} show that an LLM performs fine-grained dimensional sentiment rating at levels comparable to fine-tuned models, and can perform zero-shot  appraisal-based emotion elicitation. This suggests that structured affective processing emerges from language modeling alone.  Others benchmark emotional competence directly. Sabour et al.~\cite{sabour2024emobench} introduce EmoBench, testing emotional understanding and application, and Wang et al.~\cite{wang2023emotional} evaluate LLMs on emotional-intelligence assessments. Huang et al.~\cite{huang2024apathetic} instead measure how an LLM's reported emotional state shifts in response to situational prompts, utilizing appraisal theory. A third strand treats emotion as a functional driver. In a more adversarial setting, Long et al.~\cite{long2025evoemo} show the importance of affect. They evolve emotional policies for LLM agents engaged in multi-turn negotiation. They show that adaptive emotional expression improves negotiation performance and resists exploitation by an opposing agent, beyond what purely rational strategies achieve.

\subsection{Emotional Contagion in Crowd Simulations}
In the literature, established agent-based techniques define models based on rules and analogies to thermodynamical or epidemiological models. ESCAPES has each agent adopt a blend of its neighbors' emotions, weighted by distance and by authority figures~\cite{tsai2011escapes}. ASCRIBE models emotional intensity as
flowing between agents through coupled differential equations, where the strength of each transfer is the product of the sender's expressiveness, a channel strength, and the receiver's openness~\cite{bosse2015ascribe}.
Personality-driven crowd models adopt a generalized, epidemiological contagion mechanism that derives expressiveness
and susceptibility from OCEAN traits such as extraversion and
stability~\cite{Durupinar2016-va, basak2018using}, a family surveyed
in~\cite{vanhaeringen2023review}. Even though these parameters are drawn from
empirically validated psychological models, emotional contagion is still produced by
algorithmic rules that transfer emotion between agents. This paper bypasses such
built-in transfer rules and lets the LLM appraise the crowd instead.


\section{Method}
\begin{figure}
        \centering
        \includegraphics[width=0.69\linewidth]{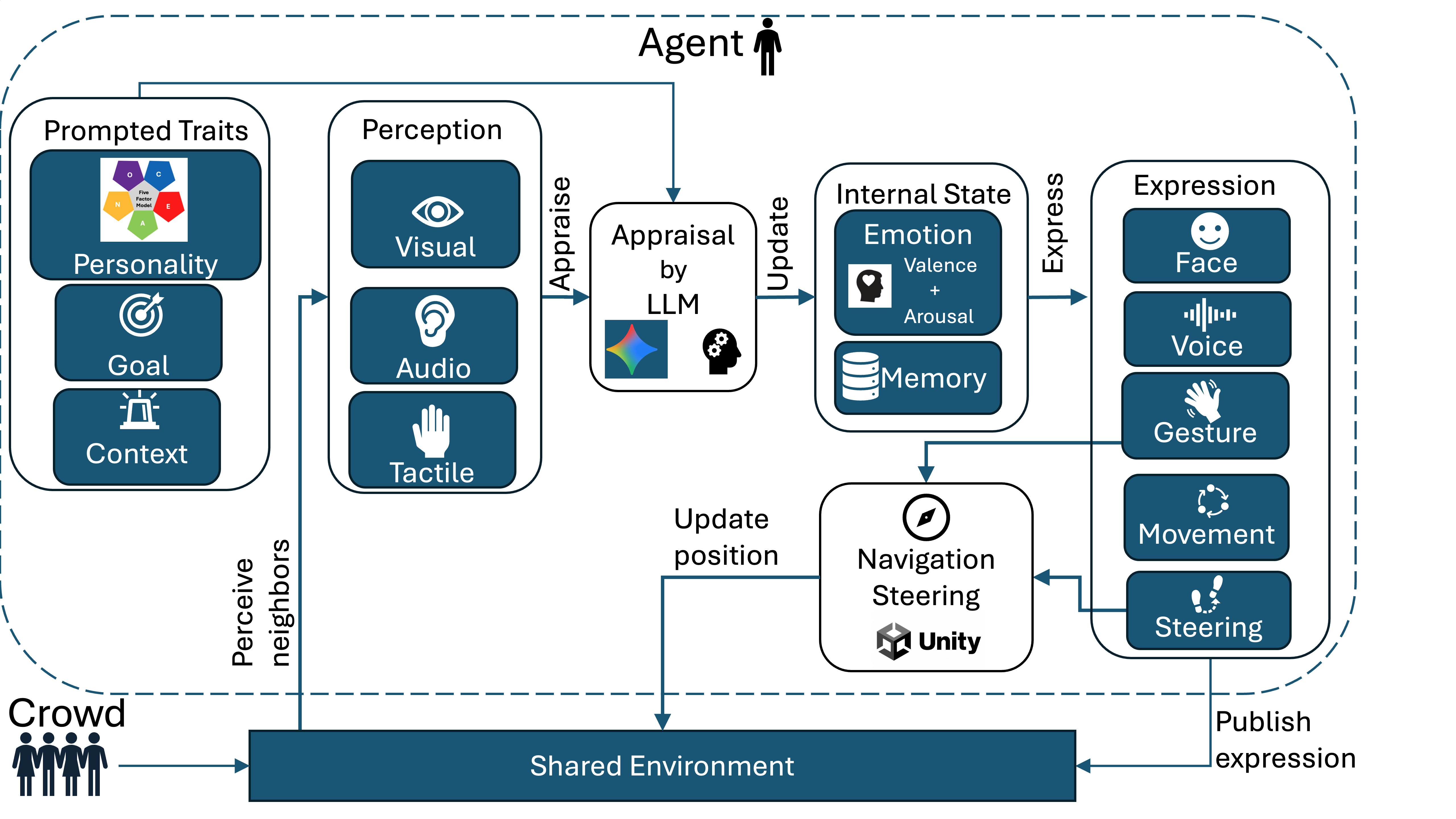}
        \caption{System architecture. Each LLM-driven agent perceives its neighbors within a shared environment, appraises their behaviors, updates its internal state, and chooses outward expression parameters. }
        \label{fig:system}    
\end{figure}

We design a system where each agent runs an LLM-controlled loop of perception $\rightarrow$ appraisal $\rightarrow$ internal state update $\rightarrow$ expression, layered on top of a conventional crowd simulation architecture in Unity~\footnote{https://unity.com}. At each step, an agent perceives the observable behavior of nearby agents, and the LLM appraises this input considering the agent's personality, current affective state, memory, and context. The appraisal updates the agent's private emotional state and its outward expression, which in turn govern how it moves through the crowd.  Figure~\ref{fig:system} presents an overview of the system architecture. We describe each component in the order that they are built on top of each other. 


\subsection{Perception}
\label{sec:perception}

Each agent observes its neighbors' external signals and computes summary
descriptors of the perceived crowd from them. Perception is split into three channels: visual, audio, and tactile.  The visual channel captures faces, gestures, movement, local density, mean speed, and motion coherence. Motion coherence is analogous to the velocity-alignment rule in Reynolds' boid model~\cite{reynolds1987flocks} and is the magnitude of the average unit heading vector. A neighbor is seen only if it lies within a limited visual range of 8~m, inside a 200\textdegree~field-of-view cone, and is not occluded by a wall or by intervening bodies. The audio channel carries vocalizations and is omnidirectional and unoccluded. However, loud calls such as screaming, shouting, laughing, and a loud reassurance call carry farther (20~m) than quiet ones such as chattering and gasping (10~m). 

In addition to vision and hearing, agents feel crowd pressure through a tactile
channel. Felt pressure measures how much the crowd holds an agent back. Taking $v_i^{\mathrm{des}}(t)$ and $v_i^{\mathrm{act}}(t)$ for agent $i$'s desired and actual speed toward its current goal, the instantaneous pressure is $\tilde p_i(t) = \big(v_i^{\mathrm{des}}(t) - v_i^{\mathrm{act}}(t)\big)/v_i^{\mathrm{des}}(t)$ clamped to $[0,1]$. It is defined to be zero whenever the agent is not pursuing a goal. This instantaneous value is smoothed over time with an exponential moving average as  $P_i(t) = \alpha\, \tilde p_i(t) + (1-\alpha)\, P_i(t - \Delta t)$, where   $\alpha \in (0,1]$. The smoothing gives more weight to recent values and lets older ones fade away gradually, reflecting a sustained pressure. Sustained high pressure can knock an agent to the ground.

To inform the LLM about the current perception, a snapshot of the perceptual state is
taken and converted into natural language by a fixed template. Formally, the perceptual
state $z_i(t) = \{x_j(t) : j \in \mathcal{N}_i(t)\}$ collects the external states of all
perceived neighbors $\mathcal{N}_i(t)$, and the converter produces the
description $D_i(t) = \mathrm{NL}(z_i(t))$ that is inserted into the LLM prompt. The description begins
with a one-line summary of the visible crowd, giving its size, density, coherence, and
mean speed, and notes any visibly fallen individuals. It then counts the visible agents displaying each facial expression, gesture, and kind of movement, and those emitting each kind of vocalization, and reports any voices that are audible but not visible, with the distance and direction of the nearest. Finally, for up to three nearest visible individuals, it gives the distance and direction together with the full expressive state, including face, gesture, voice, movement, speed, and spacing. 

\subsection{Internal State}
\label{sec:internalState}
Each agent's internal state includes a fixed personality profile and a time-varying emotional state. We represent personality using the Big Five model~\cite{goldberg2013alternative}, also known as OCEAN, which comprises openness, conscientiousness, extraversion, agreeableness, and neuroticism. Each trait is represented independently on a scale from $0$ to $1$, so agent $i$ has a fixed personality vector $\theta_i \in [0,1]^5$. We represent emotion using Russell's circumplex model~\cite{russell1980circumplex}, a two-dimensional space defined by valence and arousal. Both dimensions range from $-1$ to $1$, giving agent $i$ the time-varying emotional state $a_i(t)=\bigl(V_i(t),A_i(t)\bigr)\in[-1,1]^2$. In addition to the affective state, each agent maintains a short rolling memory $M_i(t)$ of the six most recent salient events it has perceived or performed, updated as a queue: $M_i(t+\Delta t) = \big(M_i(t)\big)_{[-(K-1):]} \,\Vert\, m_i^{\mathrm{new}}$, where the oldest note is dropped once the queue exceeds length $K$ and the newest note $m_i^{\mathrm{new}}$ is appended. We take $K=6$, which corresponds to approximately the previous $15$--$20$ seconds of the simulation when notes accrue at every appraisal. This window provides short-term temporal context without substantially increasing the prompt length. Memory entries are stored as text so that they can be inserted directly into the agent's next LLM prompt.

\subsection{External State}
\label{sec:externalState}
The external state comprises the agent's expressive behaviors, which the LLM selects as part of its appraisal output. It is formulated as
 \[
 x_i(t) = \big(F_i(t), G_i(t), \mathit{Vo}_i(t), \mathit{Mv}_i(t), \sigma_i(t), \rho_i^{\mathrm{space}}(t), \phi_i(t), n_i(t)\big)
 \]
These include a facial expression $F_i$, a gesture $G_i$, a vocalization $\mathit{Vo}_i$, a movement choice $\mathit{Mv}_i$, and a set of steering parameters. The facial expressions are drawn from Ekman's basic emotions~\cite{ekman1971constants} and consist of joy (represented as a smile), fear, anger, surprise, sadness (represented as a frown) together with a neutral face. Gestures include waving, beckoning, shielding, pointing in alarm, cheering, dancing, and slumping, or none. Vocalizations range over silence, chatter, laughter, shouting, screaming, gasping, and a loud reassuring call. The movement style can be idle, strolling, purposeful, agitated, or fleeing. Finally, the steering parameters tie these choices to the crowd simulation and consist of the personal space the agent tries to keep ($\rho_i^{\mathrm{space}}$), a speed multiplier ($\sigma_i$), a forcefulness value ($\phi_i$), and a navigation intent ($n_i$) of proceeding, halting, following a leader, or helping another agent. All components of $x_i(t)$ are set at each appraisal time and held fixed for perception and locomotion until the next appraisal.

\subsection{Appraisal by LLM}
\label{sec:appraisalLLM}
The LLM serves as the coordinator of an agent's internal state and the selector of its external expression. Unlike prior approaches in which psychological states are mapped to steering parameters and behavioral choices through author-defined rules~\cite{durupinar2009ocean, Durupinar2016-va}, here the mapping from state to behavior is produced by the LLM. At each step the agent is prompted to update its emotional state from its appraisal of the situation and its personality. The fixed system prompt is provided in Appendix~\ref{app:prompt}. Besides this, on each appraisal step, the agent constructs a prompt containing the following information.
\begin{itemize}[leftmargin=*, itemsep=0pt]
    \item[-] Identity and context: its personality, goal, current action, and situational context (e.g., a concert).
    \item[-] Affective state: its current valence and arousal.
    \item[-] Memory: a short rolling memory of salient recent events.
    \item[-] Perception: a natural-language summary of what it currently sees and hears, together with the crowd pressure it feels.
\end{itemize}

Formally, at each appraisal time $t$ the LLM $f_{\mathrm{LLM}}$ returns a single JSON
object comprising a target affect, an expression choice for each communication modality, a
brief private rationale, and a one-line memory note:
\footnotesize
\[
\big(V_i^{\mathrm{tgt}},\, A_i^{\mathrm{tgt}},\, x_i^{\mathrm{new}},\, r_i,\,
m_i^{\mathrm{new}}\big)
= f_{\mathrm{LLM}}\big(\theta_i,\, a_i(t),\, x_i(t),\, M_i(t),\, D_i(t),\, P_i(t),\,
c_i\big),
\]
\normalsize
where $r_i$ is the private reasoning string, $m_i^{\mathrm{new}}$ the memory note appended
to $M_i$, and $c_i$ the scenario context and the agent's goal. The affective state is then
updated by linearly interpolating from the current state toward the target, applied only
at appraisal times:
\[
d_i(t+\Delta t) = (1-\lambda)d_i(t) + \lambda d_i^{\mathrm{tgt}},
\]
where  $d\in \{V,A\}$ and $\lambda \in (0, 1]$ is a smoothing factor. We take
$\lambda = 0.5$ in our simulations (sensitivity analysis is provided in Appendix~\ref{app:sensitivityLambda}).

Appraisal runs on a per-agent staggered clock. Agent $i$ nominally re-appraises at times
$t \in \{t_i^{(0)}, t_i^{(0)} + T, t_i^{(0)} + 2T, \ldots\}$, with period $T = 3$\,s and
an agent-specific phase offset $t_i^{(0)}$ that staggers calls across the crowd (sensitivity analysis for $T$ is provided in Appendix~\ref{app:sensitivityT}). Because
each agent keeps its own timer, the interval does not depend on how many others are
present, provided the backend sustains the aggregate request rate. A global limit bounds
the number of simultaneous LLM calls, and at our crowd sizes the achieved median interval
is $3.5$\,s across all runs. The backend is pluggable and can issue calls to an external API or to a local server. Movement and the expression of observable signals are decoupled from this slow cognitive clock and updated every frame.

\subsection{Locomotion and Steering}
\label{sec:locomotion}
Low-level navigation is decoupled from the LLM and handled conventionally by the Unity crowd simulator. We keep locomotion separate from LLM decisions as real-time
geometric path planning and obstacle avoidance require precise, continuous control that is
beyond the reliable capability of a zero-shot LLM. Additionally, locomotion must update every
frame, whereas appraisal runs at a much lower frequency. Having the LLM handle steering would require a call per agent per frame, which is prohibitive in latency and cost.

In Unity, local collision avoidance uses reciprocal velocity obstacles~\cite{van2008reciprocal} and global path planning uses A* over a navigation mesh. Since shortest paths tend to pull agents into a single file, we add randomized lateral spreading, giving each agent a small gait variation, a random
avoidance priority, and a fixed lane preference that drifts it toward one side of its
route as far as free space allows. Thus, the crowd  fans into parallel lanes in open
rooms and tucks into a tight column where a corridor narrows. A separation force nudges
agents apart without pushing them through walls.

Emotion enters the motion only through the appraisal outputs. The chosen movement style
selects the manner of moving, from idle to fleeing. The navigation intent, drawn from
\{Proceed, FollowLeader, Halt, Help\}, selects the current objective, respectively
proceeding toward one's own goal, moving with a nearby calm and beckoning agent, freezing
in place, and going to a nearby agent that looks panicked, frozen, or has fallen. Three
continuous parameters then shape execution, with the speed multiplier scaling walking
pace, forcefulness scaling how hard the agent presses into others, and the personal-space
multiplier scaling the clearance it keeps. Sustained crowd pressure can additionally knock an agent to the ground. A fallen agent is an observable signal that neighbors may respond to through the Help intent, though falls are rare in the low-density scenarios we studied.

\section{Evaluation}
\subsection{Setup}
We describe a group of scenarios created to validate the system and test our hypotheses. A scenario is a data object that specifies the crowd size, spawn region(s), goals, walls and obstacles, seeding, and a personality distribution profile.  Agent counts are set to control the local density each agent perceives, which we choose per scenario.  Goals are defined as scatter zones, letting each agent sample its own target within a region. Seeding determines which agents act as triggers and their affective states. For instance, agents can be seeded with certain valence and arousal values or personality traits to make them act alarmed, angry, or elated. 

Each agent's five personality traits are drawn independently from normal distributions clamped to  $[0,1]$. A baseline crowd always uses $\mu = 0.5$ and $\sigma = 0.16$ for each trait, providing coverage across the trait range without causing values to accumulate at the bounds. All sampling uses fixed random number generator seeds (different from the affective seeding above), to ensure that every crowd and run is reproducible.  

\begin{figure*}[t]
    \centering

    \begin{subfigure}[b]{0.32\linewidth}
        \centering
        \includegraphics[width=\linewidth]{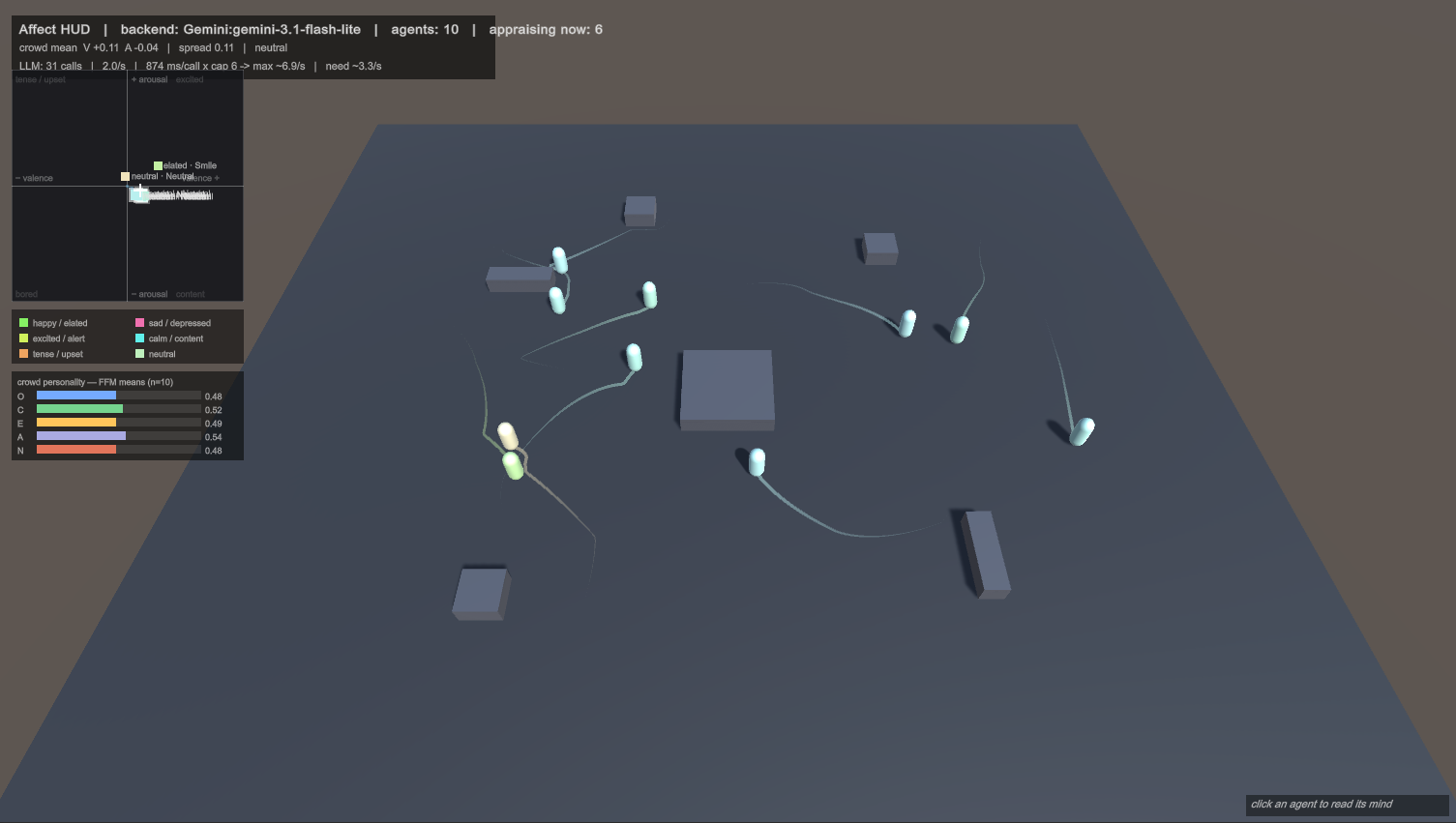}
        \caption{Calm Plaza}
        \label{fig:calmPlaza}
    \end{subfigure}
    \hfill
    \begin{subfigure}[b]{0.32\linewidth}
        \centering
        \includegraphics[width=\linewidth]{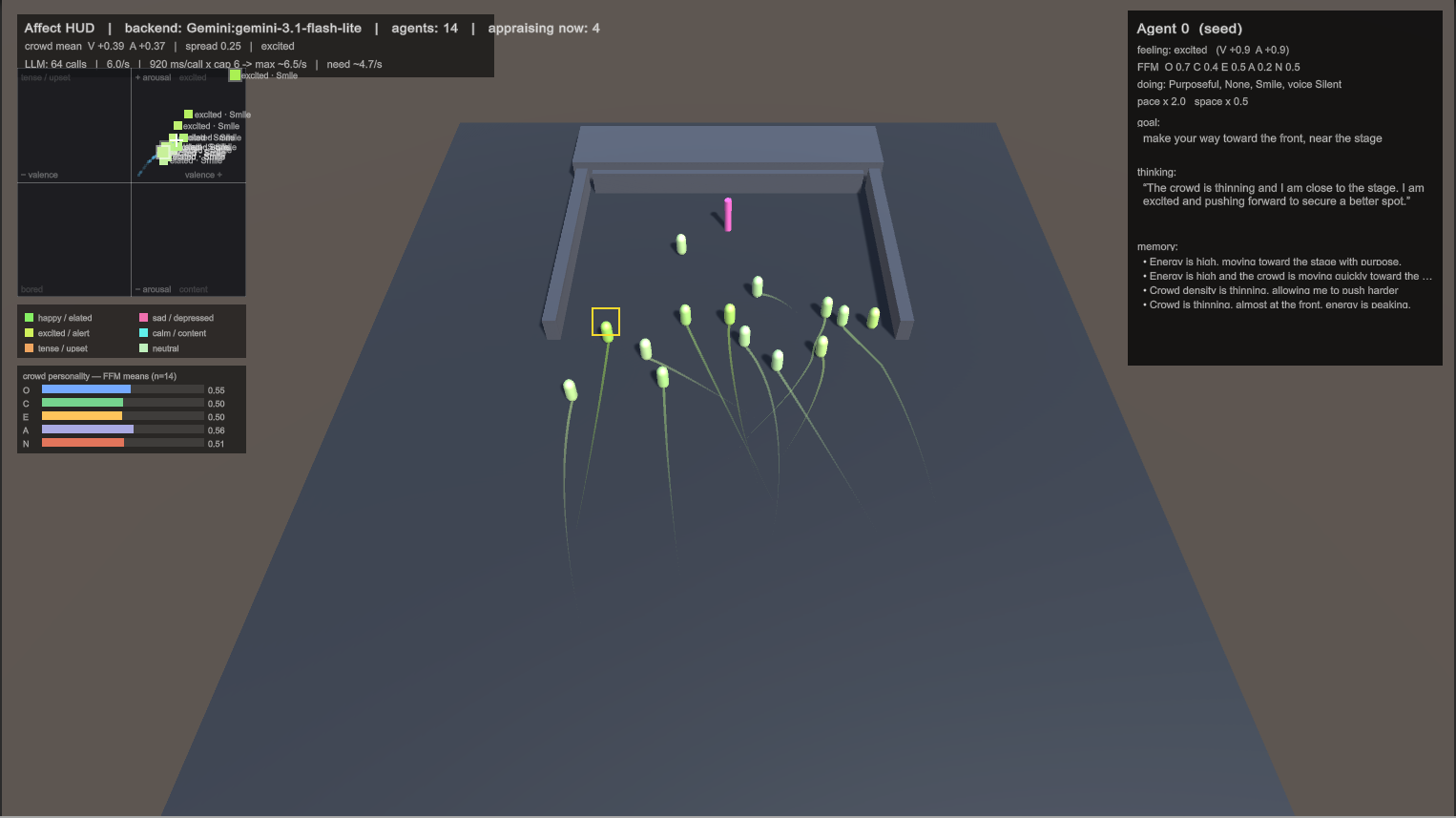}
        \caption{Concert}
        \label{fig:concert}
    \end{subfigure}
    \hfill
    \begin{subfigure}[b]{0.32\linewidth}
        \centering
        \includegraphics[width=\linewidth]{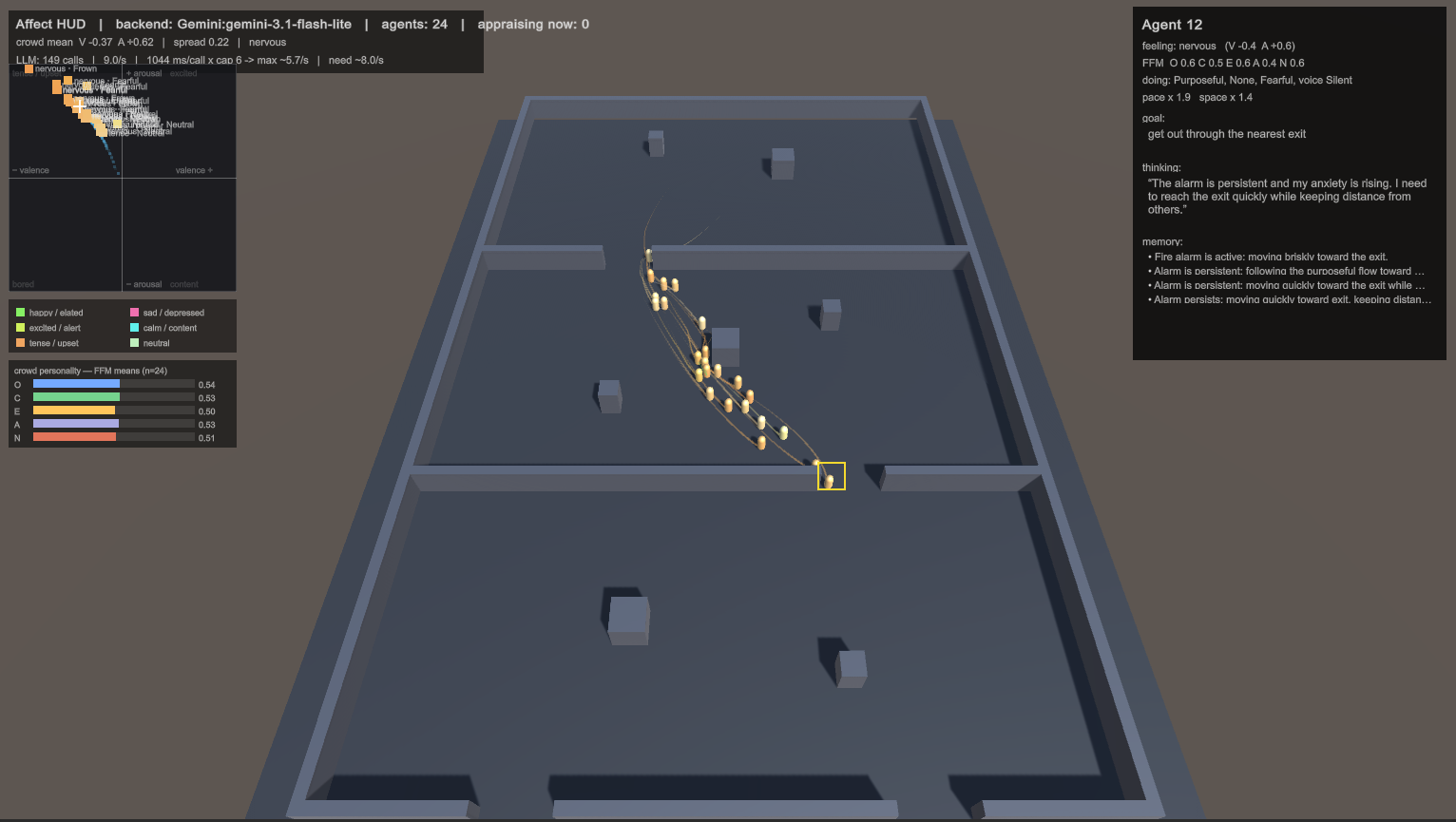}
        \caption{Evacuation}
        \label{fig:evacuation}
    \end{subfigure}

    \vspace{0.5em}

    \hfill
    \begin{subfigure}[b]{0.32\linewidth}
        \centering
        \includegraphics[width=\linewidth]{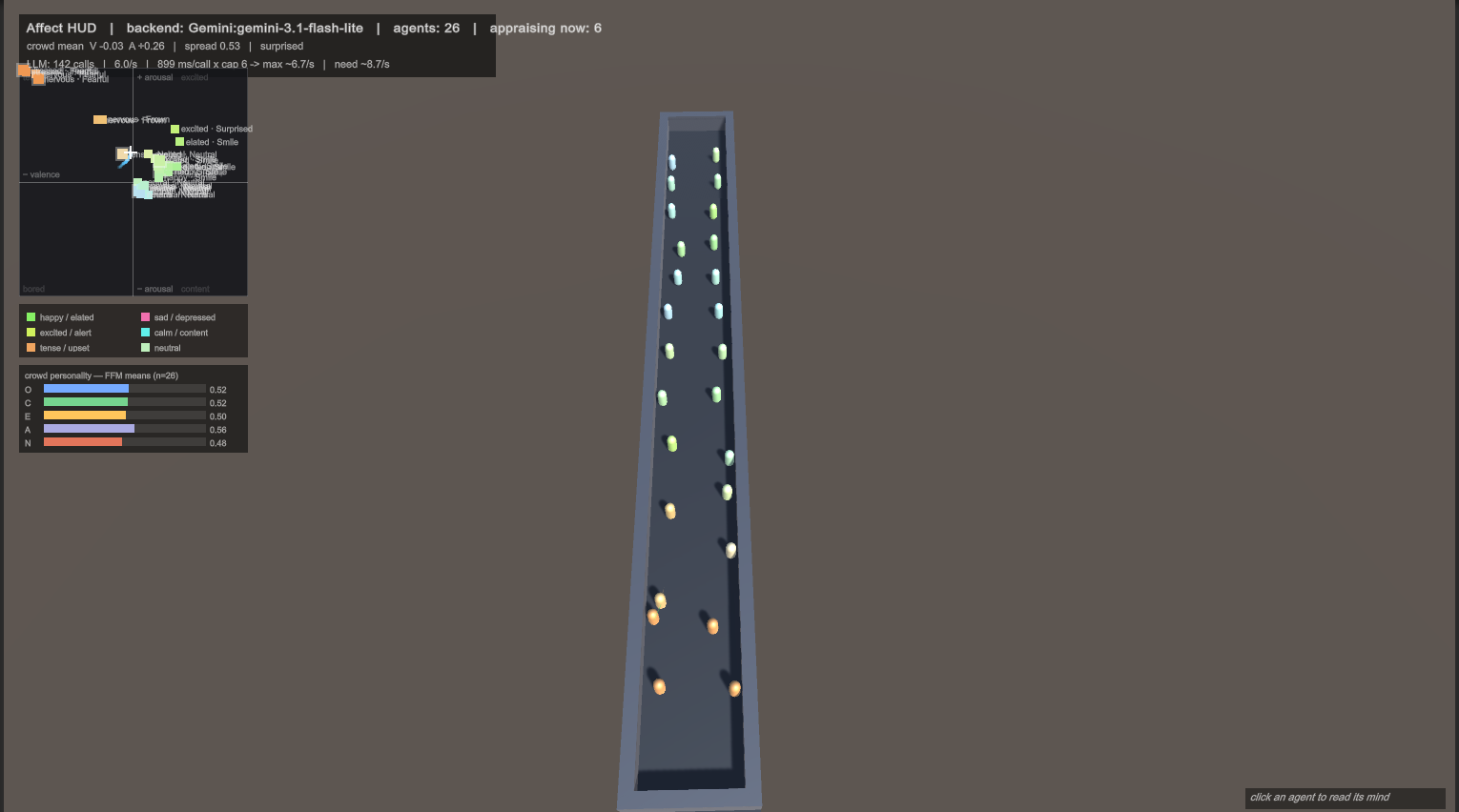}
        \caption{Standing Line}
        \label{fig:standingLine}
    \end{subfigure}
    \hfill
    \begin{subfigure}[b]{0.32\linewidth}
        \centering
        \includegraphics[width=\linewidth]{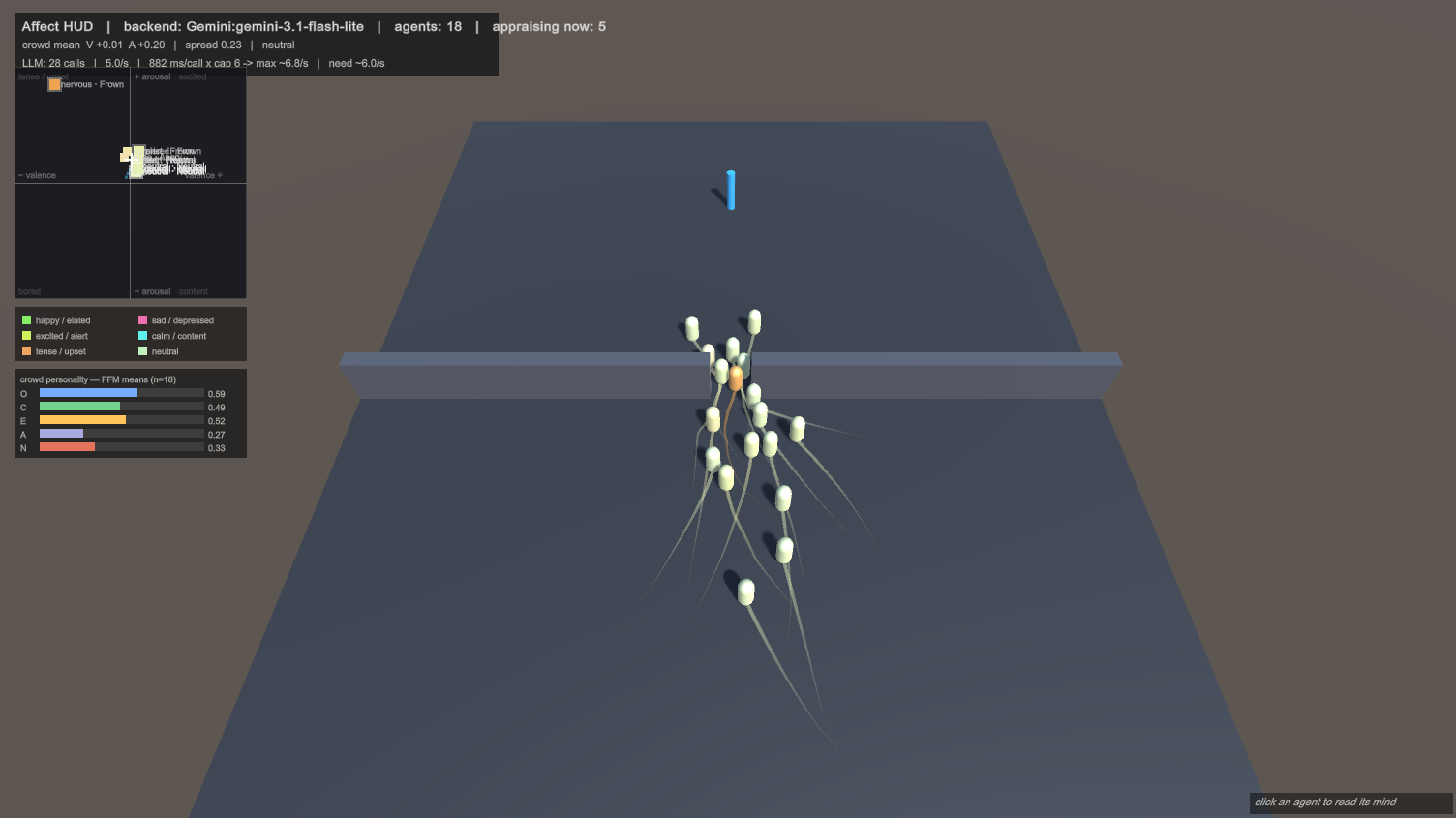}
        \caption{Contested Gate}
        \label{fig:contestedGate}
    \end{subfigure}

    \caption{Simulation scenarios used in our experiments. Agents are colored by their
affective state on the circumplex: hue gives the direction of the feeling (green
pleasant, yellow activated, red unpleasant, blue deactivated) and saturation its
intensity, so near-neutral agents appear pale. In the interactive simulator,
selecting an agent reveals its internal state and appraisal reasoning. }
    \label{fig:scenarios}
\end{figure*}

Screenshots from the scenarios are displayed in Figure~\ref{fig:scenarios}. Their descriptions are as follows. 

\begin{itemize}[leftmargin=0pt, topsep=1pt, partopsep=0pt, itemsep=0pt]
    \item[]\textbf{Calm Plaza.} An open $56 \times 56$\,m room where 10 agents stroll between points of interest  
    (Figure~\ref{fig:calmPlaza}). It tests baseline behavior, where affective state should stay near neutral. 
    \item[]\textbf{Concert.} An open $54 \times 70$\,m room with a stage separated from a wide scatter zone by a pit rail (Figure~\ref{fig:concert}). 14 agents are spawned. It tests positive contagion for excitement and joy.
    \item[]\textbf{Evacuation.} A multi-room $70 \times 108$\,m building with staggered internal doorways and two far exits (Figure~\ref{fig:evacuation}). 24 agents flee toward the exits. Depending on the test, agents are either removed when they reach the exit or gather in a safe zone after exiting. It tests panic spread and calming down.
    \item[]\textbf{Standing Line.} A long, narrow $9 \times 80$\,m lane. 26 agents mill in place with no context, with a panic cluster seeded at one end (Figure~\ref{fig:standingLine}). It tests spatial propagation of panic as a traveling wave-like front.
   \item[]\textbf{Contested Gate.} An open $44 \times 60$\,m area split by a wall with a single 2.4\,m gate (Figure~\ref{fig:contestedGate}). 18 agents spawn on one side and must reach the other, passing through the gate. It tests anger contagion in the presence of seeded provocateurs.
    
\end{itemize}

The exact prompts and a summary table for the scenarios are provided in Appendix~\ref{app:scenarios}.
Agent counts are selected considering the neighbors each agent perceives. Based on the scene geometry, the numbers are bounded below by the need for a few co-perceiving neighbors to ensure contagion. They are bounded above by the LLM appraisal budget, since every agent consumes a call each step. The mean number of visually perceived neighbors within the $8$\,m visual range is $\approx 3$ for Concert and Evacuation, $\approx 2$ in Standing Line, and $\approx 1$ for the Calm Plaza control.

In our experiments we use the  \texttt{gemini-3.1-flash-lite}~\footnote{Gemini via Google Generative Language API (\texttt{v1beta}, \texttt{generateContent}); JSON output mode, $200$ max tokens, 0.4 temperature, other parameters left at default values,  June--July 2026.} model as the main LLM for a balance of speed and cost. We call the API with a temperature of 0.4 to balance determinism with a little creative variance when the LLM appraises the given description of the internal state. 

Each experiment evaluates a separate hypothesis. Therefore, we apply Holm--Bonferroni correction within each hypothesis, controlling the family-wise error rate.  If a hypothesis has sub-hypotheses tested in different scenarios, such as H1 and H5, each sub-hypothesis forms its own family. A family comprises the tests that directly evaluate that hypothesis or sub-hypothesis; robustness fits that re-examine an already-counted coefficient under an alternative specification, such as adding a distance-to-seed covariate or run fixed effects, are not counted as additional tests. Reported $p$ values are Holm--Bonferroni adjusted within each family.

\subsection{Experiments}

\subsubsection{Local Emotional Contagion}~\label{sec:localEmotionContagion}

\noindent\textbf{Design.} This experiment is designed to test \textbf{H1} within Evacuation, Concert, and Calm Plaza scenarios. We run each scenario 20 times with a different random draw for a baseline personality distribution.  Each scenario tests a different case. For Evacuation, where a single agent is seeded to be panicked ($V = -0.85$, $A = 0.85$), the expectation is that neighbors' observable alarm signals cause a drop in valence and an increase in arousal.  For the Concert, where  a single agent is seeded to be elated ($V = 0.85$, $A = 0.85$), we expect the observable positive expressions of neighbors to drive an increase in valence and arousal. The Calm Plaza setting is the negative control.

\minititle{Analysis.} For each scenario we report the crowd's mean valence and arousal, their trajectories over normalized run time, the fraction of perceived neighbors displaying alarm or positive cues, a run-level cross-sectional comparison, and a lagged regression analysis. The cross-sectional comparison tests whether agents surrounded by alarmed (in Evacuation) or positive (in Concert) neighbors are themselves more aroused and more negative (in Evacuation) or more positive (in Concert) than agents in neutral surroundings.  Alarm cues are fearful faces, fleeing motion, screaming/gasping, shielding/pointing-alarm gesture, or a fallen body. Positive cues include smiling, laughing, and cheering/dancing.   For each run we compute the mean affect over appraisal steps in alarmed (or positive) neighborhoods and over steps in neutral ones, then compare the two within-run means with a paired t-test across runs.  

For regression analysis, we fit two linear models, one for the change in valence and one for the change in arousal for the non-seeded agents. For this lagged test, the neighbor cues are read at the current step $t$ and the agent's affect change is predicted at the next step $t+1$, controlling for the agent's own state and local density. For each agent $i$ and run $r$, we compute the change in valence, $\Delta V$, and the change in arousal, $\Delta A$, from time $t$ to $t+1$. The model is:
\footnotesize
\[
\begin{aligned}
\Delta d_{i, t\rightarrow t+1, r}
&=
\beta_0^{d}
+ \beta_1^{d}\,\text{neighbor alarm}_{i, t, r}
+ \beta_2^{d}\,\text{neighbor positivity}_{i, t, r} \\
&\quad
+ \beta_3^{d}\,\text{local density}_{i, t, r}
+ \beta_4^{d}\,\text{own valence}_{i, t, r}
+ \beta_5^{d}\,\text{own arousal}_{i, t, r}\\
&\quad
+ \varepsilon^{d}_{i,t,r},  
\end{aligned}
\]
\normalsize
where $d \in \{V, A\}$ and $\beta^{d}_j$, $j \in \{0,\dots,5\}$, are the regression coefficients.
Standard errors are clustered by run, which makes no within-run independence assumption.   For Evacuation, we expect $\beta^{V}_1 < 0$ and $\beta^{A}_1 > 0$, and for Concert, we expect $\beta^{V}_2 > 0$ and $\beta^{A}_2 > 0$.

\minititle{Results.} In Evacuation, $39\%$ of the appraisal logs show alarm cues; mean
valence is $-0.32$, and mean arousal is $0.56$. Averaged across the runs, arousal climbs
from $0.24$ to $0.69$ and valence falls from $-0.12$ to $-0.35$, while the neighbor-alarm
fraction rises from $0.06$ to $0.20$. A cross-sectional comparison shows that agents among
alarmed neighbors are more aroused and more negative than those in calm neighborhoods,
with arousal $0.62$ vs.\ $0.44$ ($t = 5.8$, $p < 0.001$, $n=20$) and valence $-0.37$
vs.\ $-0.22$ ($t = -5.8$, $p < 0.001$, $n = 20$). The lagged regression points the same
way. Neighbor alarm at the current step predicts a next-step valence drop,
$\beta^{V}_1 = -0.030$ ($t = -5.0$, $p < 0.001$, 95\% CI $[-0.042, -0.019]$), and an
arousal rise, $\beta^{A}_1 = 0.026$ ($t = 4.3$, $p < 0.001$, 95\% CI $[0.014, 0.037]$),
controlling for own state and density, with $n = 4997$ agent-steps across 20 runs. Adding
each agent's distance to the seed as a covariate leaves the neighbor-alarm coefficients
unchanged, with distance itself contributing negligibly ($|\beta| \le 0.0003$ per
meter); thus, the effect is not a proximity artifact. However, Evacuation is a short
simulation ($\approx 50$--$60$\,s per run), where agents are removed as they reach the exit.
When run fixed effects and a within-run time trend are added, estimating the effect only
from variation within each run, both neighbor-alarm coefficients are indistinguishable
from zero ($\beta^{V}_1 = 0.002$, $p = 0.76$; $\beta^{A}_1 = -0.008$, $p = 0.31$).
We therefore base H1.a on the trajectory and the run-level cross section. Although the
lagged coefficient supports the same conclusion directionally, it cannot cleanly separate
step-to-step contagion from run-level differences.

In Concert, $71\%$ of appraisals show positive cues, mean valence is $0.39$, and mean
arousal is $0.56$. Averaged across the runs, valence climbs from $0.11$ to $0.52$ and
arousal from $0.16$ to $0.83$, as neighbor positivity rises from $0.21$ to $0.77$.
Cross-sectionally, agents in positive neighborhoods have much higher valence, $0.50$
vs.\ $0.08$ ($t = 19.1$, $p < 0.001$, $n=20$), and higher arousal, $0.55$ vs.\ $0.31$
($t = 3.9$, $p = 0.003$, $n=20$). For the lagged test, neighbor positivity predicts a
next-step valence rise, $\beta^{V}_2 = 0.033$ ($t = 2.1$, $p = 0.035$, 95\% CI
$[0.002, 0.063]$), $n = 4147$ across 20 runs. The arousal coefficient is small and of the
opposite sign, $\beta^{A}_2 = -0.011$ ($p = 0.005$, $[-0.018, -0.004]$). Thus, the crowd's
rising arousal is driven by the setting and not a lagged neighbor-arousal transfer,
supporting H1.b only for valence.

\begin{figure*}
    \centering
    \includegraphics[width=0.8\linewidth]{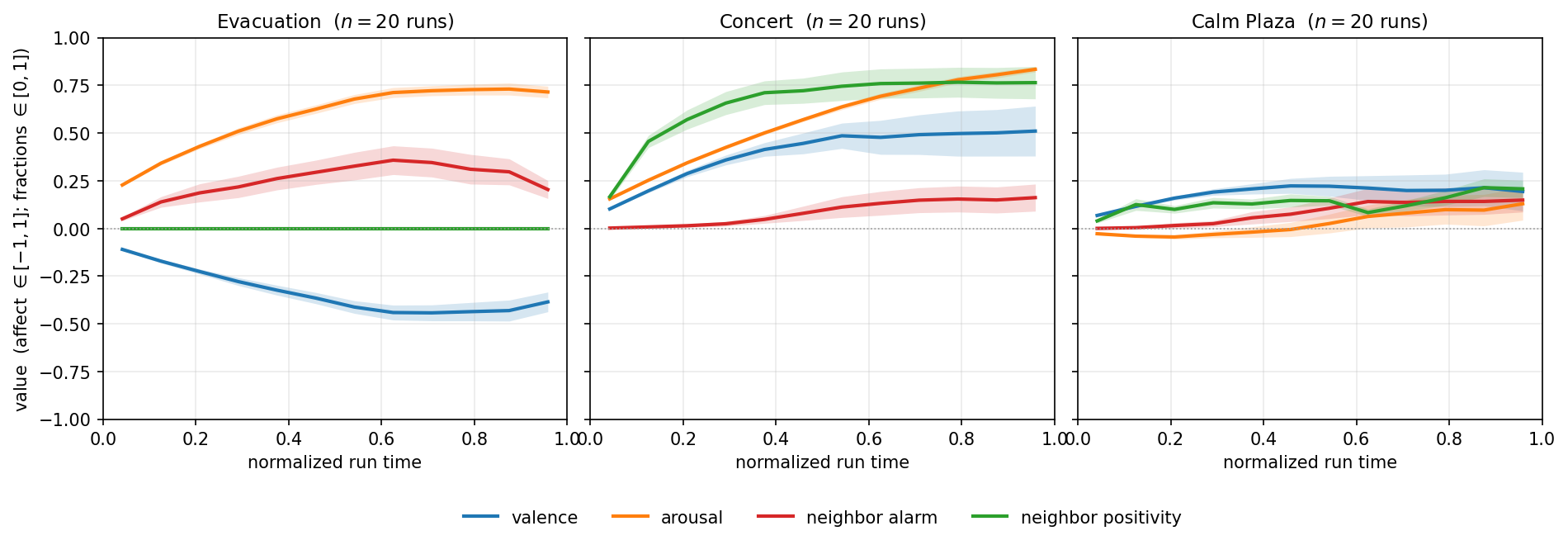}
    \caption{Affect trajectories across three scenarios.  \textbf{Left:} Evacuation builds into
    crowd-wide panic from a single seed. \textbf{Center:} Concert is a positive cascade from a single
    seed. \textbf{Right:} Calm Plaza stays mildly pleasant, never tipping into panic.}
    \label{fig:h1}
\end{figure*}

For the Calm Plaza, without a seed and in an uncrowded space, only $11\%$ of appraisals
show alarm cues and $23\%$ positive cues, mean valence is $0.18$, and mean arousal is
$0.03$. Averaged across the runs, valence and arousal stay near neutral, valence changing
from $0.07$ to $0.19$ and arousal from $-0.03$ to $0.13$. The neighbor-alarm fraction
rises from $0.00$ to $0.15$ and neighbor positivity from $0.05$ to $0.21$.
Cross-sectionally, agents in alarmed neighborhoods are more negative (valence $-0.17$
vs.\ $0.16$, $t = -3.8$, $p = 0.005$) and more aroused ($0.41$ vs.\ $0.00$, $t = 4.6$,
$p = 0.004$) than those in calm neighborhoods, though only 9 of the 20 runs contain any
alarmed neighborhood. In the rare case where an agent sees an alarmed neighbor, that alarm
predicts a next-step valence drop and arousal rise, $\beta^{V}_1 = -0.095$ ($p < 0.001$,
95\% CI $[-0.137, -0.054]$) and $\beta^{A}_1 = 0.078$ ($p < 0.001$, $[0.044, 0.112]$),
$n = 2111$ across 20 runs; and unlike Evacuation this coupling stays stable within runs,
holding under run fixed effects and a within-run time trend ($\beta^{V}_1 = -0.074$,
$p = 0.006$; $\beta^{A}_1 = 0.072$, $p = 0.005$). However, the overall crowd stays calm as
such cues are scarce. With almost nothing alarming to catch, the crowd's mood is set by
its direct appraisal of the benign setting, and it stays mildly pleasant without tipping
into panic, consistent with H1.c at the crowd level. Figure~\ref{fig:h1} shows the crowd's average valence, arousal, neighbor-alarm and neighbor-positivity fractions over normalized run time across the three scenarios.

\subsubsection{Spatial Propagation}~\label{sec:wave}

\noindent\textbf{Design.} This experiment tests \textbf{H2}, which asks whether panic from a localized source has a spatial signature in the form of a traveling wave. We use the Standing Line scenario, where agents mill in place, occasionally turning and glancing around to observe their neighbors. A small panic cluster of four agents ($V = -0.85$, $A = 0.85$) is seeded at one end. The seeds are given high neuroticism ($0.9$) to make sure they consistently express their panic. The scene involves a long and thin lane to isolate each agent's coupling only to nearby line neighbors.  We run the scenario 30 times, each with a different baseline crowd.
 
\minititle{Analysis.}  For each non-seed agent, we record its initial distance to the seed cluster and the first time at which it enters the panic quadrant with $V < 0$ and $A > 0.35$ (see sensitivity analysis in Appendix~\ref{app:sensitivityPanic}). Among agents that enter the panic quadrant, we regress this onset time on initial distance from the nearest seed. A positive slope indicates that more distant agents tend to enter the panic quadrant later, consistent with outward propagation. Because the slope has units of seconds per meter, its inverse provides a descriptive propagation-speed estimate in ($\mathrm{m/s}$), provided that the relationship is approximately linear.  To assess the reliability of this relationship across runs, we calculate the Pearson correlation between initial distance and onset time separately within each run. We apply the Fisher $z$-transformation to the run-level correlations and use a one-sample t-test to determine whether their mean differs from zero across the 30 runs.


Since the correlation and speed are computed only over the agents that entered the panic quadrant and because the wave does not reach the whole crowd, these measures describe propagation among affected agents. To assess the spread of panic through the full crowd, our main measure is the steady-state affect gradient. We use the final $25\%$ of each run as a late-run analysis window. Within this window, we calculate each agent's mean arousal, mean valence, and proportion of appraisal steps spent in the panic quadrant. For each outcome, we calculate its Pearson correlation with the agent's initial distance from the nearest seed separately within each run. The resulting correlations are Fisher-$z$-transformed and tested against zero across the 30 runs using one-sample t-tests. These gradients indicate how affect is distributed relative to the seed late in the run, by which point the alarmed fraction has plateaued. A negative correlation between distance and arousal or panic-quadrant occupancy indicates that alarm remains concentrated near the seed. A positive correlation between distance and valence indicates that agents farther from the seed remain more positive. Correlations near zero indicate no linear distance gradient, although they do not necessarily establish that affect is uniformly distributed.

\minititle{Results.} The seeded cluster shows alarm cues (fearful faces, gasps/screams,
shielding/pointing) on $98\%$ of its appraisals. Across the 30 runs, the mean within-run
correlation between initial distance and panic-onset time is $r = 0.65$ and positive (one-sample $t = 5.1$, $p < 0.001$). Among the 278 agents that enter the panic
quadrant, the pooled correlation between initial distance and onset time is $r = 0.44$
($p < 0.001$). The estimated slope is $1.04$~s/m, with a run-level bootstrap 95\% CI
$[0.67, 1.53]$~s/m. Inverting the slope gives a descriptive propagation-speed estimate of
$\approx 0.96$~m/s (95\% CI $[0.65, 1.49]$~m/s).

The steady-state gradients show that alarm remains spatially concentrated near the source.
Mean arousal decreases with initial distance from the seed, with a mean within-run
correlation of $r = -0.49$ ($t = -7.2$, $p < 0.001$).  The proportion of time spent in the panic quadrant also decreases with distance, $r = -0.56$ ($t = -10.2$, $p < 0.001$), whereas mean valence increases with distance, $r = 0.57$ ($t = 9.1$, $p < 0.001$).

\begin{figure*}
    \centering
    \includegraphics[width=0.95\linewidth]{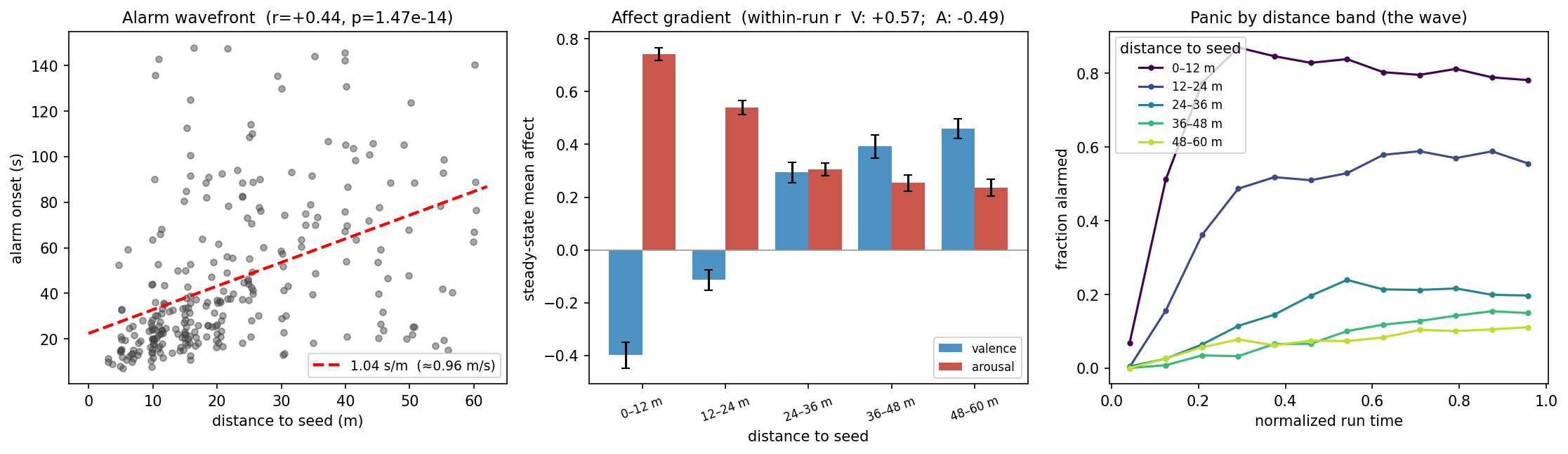}
\caption{Panic propagates outward from the seed cluster (Standing Line). \textbf{Left:}
alarm onset versus each agent's distance to the seed cluster; the fitted slope is positive
at $1.04$~s/m. \textbf{Center:} steady-state mean valence (blue) and arousal (red) by
distance band. \textbf{Right:} fraction of appraisals alarmed over normalized run time,
split by distance band.}
    \label{fig:spatialWave}
\end{figure*}

Thus, alarm remains strongest near the seed, and does not become uniformly distributed
throughout the crowd. The wave spatially reaches on average $63\%$ of the line, ranging from $19\%$
to $100\%$ across runs, so panic does not consistently propagate across the full line. Agents nearer the seed cluster are also more likely to catch the alarm at all. Within $12$\,m, $97\%$ of agents enter the panic quadrant at some point during the run, falling to $69\%$, $31\%$, $21\%$, and $15\%$ across successive $12$\,m bands.  Figure~\ref{fig:spatialWave} summarizes the  results.
The positive relationship between distance and onset time and the persistent steady-state affective gradients are consistent with outward propagation from the seeded cluster and support H2.


\subsubsection{Comparison with Epidemiological Contagion Models}
\label{sec:sir}
\noindent\textbf{Design.} This experiment tests \textbf{H3}, which asks whether the
emergent macro dynamics follow a classical epidemic model. The natural analogue is
alarm spreading from a seeded agent. Since epidemic models assume a well-mixed population we use the Evacuation setup where agents move freely. For comparison, we use the
SIR and SIS
models~\cite{hethcote2000mathematics}, each with a transmission rate $\beta$ and a
recovery rate $\gamma$.   Because personality-based crowd-contagion models such as Durupinar
et al.~\cite{Durupinar2016-va} build on the Dodds--Watts threshold
contagion~\cite{dodds2005generalized}, we also test whether the macro-level response curve exhibits a threshold pattern consistent with the existence of a critical mass required for contagion to occur. Since an epidemic comparison requires a recovery regime, we
modify the Evacuation scenario, where agents are led to a safe outdoor assembly zone
instead of removing them when they exit. For this experiment, we take 20 runs with
different baseline crowds. To observe recovery, we keep the simulation running for 180 seconds as
the agents wait in the assembly zone.

\noindent\textbf{Analysis.}
For each of the 20 runs, we calculate the fraction of alarmed agents, $I(t)$, in fixed
4-second time bins. An agent is classified as alarmed when it occupies the panic quadrant
($V < 0$, $A > 0.35$). We fit SIR and
SIS models to the observed $I(t)$ trajectory using
least squares. The initial infected fraction, $I(0)$, is fixed at the alarmed fraction
observed in the first bin, while the transmission rate, $\beta$, and recovery rate,
$\gamma$, are estimated. For each model, we report the root mean squared error (RMSE)
between the fitted and observed trajectories, the basic reproduction number
$R_0 = \beta/\gamma$, and the model's long-run
alarmed fraction.

The models differ in their long-run behavior. In the SIR model, the infected fraction
eventually approaches zero as agents recover and become resistant to reinfection. In the
SIS model, recovered agents return to the susceptible state, allowing the alarmed fraction
to approach a nonzero endemic equilibrium, $ I^{*} = 1 - \frac{\gamma}{\beta}$ when $\beta > \gamma$.

Both SIR and SIS represent simple contagion, in which exposures contribute independently
to transmission. To test for nonlinear or threshold-like contagion, we extend the SIS
model by replacing the standard transmission term $\beta I(1-I)$ with $ \beta I^{h}(1-I). $
The case $h = 1$ recovers the standard SIS model, whereas $h > 1$ indicates that
transmission increases disproportionately with the alarmed fraction and may produce
critical-mass behavior.

\noindent\textbf{Results.}
Averaged across the 20 runs, the alarmed fraction increases from $5\%$ to a peak of $47\%$ within the first $20$ s. It then declines as the crowd reaches the safe assembly area and stabilizes at a nonzero plateau of $22\%$ ($95\%$ CI $[0.12, 0.31]$). However, the plateau varies substantially across runs, with individual-run late-stage levels ranging from $0.00$ to $0.76$. Therefore, instead of its precise level, we treat the persistence of a nonzero plateau as the main finding.  This persistent alarm level is qualitatively consistent with an SIS-like process. A least-squares SIS model reproduces the pooled trajectory with an $\mathrm{RMSE}$ of $0.076$ and a positive estimated recovery rate. When the SIS model is fitted separately to each run, the transmission and recovery rates are poorly identified individually, but their ratio is comparatively stable. Nineteen of the 20 runs yield $R_0>1$. The median $R_0$ is $1.4$, with an interquartile range from $1.2$ to $1.5$. The corresponding endemic equilibrium $I^*$ has a median of $0.27$, with an interquartile range from $0.18$ to $0.34$.  The persistent plateau is consistent with the absence of immunity in the simulation. Agents who leave the panic state remain susceptible and may become alarmed again as they continue to appraise panicked neighbors.

The SIR model provides a poorer fit to the pooled trajectory, with an $\mathrm{RMSE}$ of $0.128$. By construction, with fixed $\beta$ and $\gamma$ the infected fraction must decay to zero. To compensate for the sustained plateau in our data,  the fit drives the recovery rate to near zero
($\gamma = 0.02$), which slows the decay enough to track the plateau within the run but
inflates $R_0$ to an uninterpretable $5.0$.

The threshold-based generalized SIS model does not improve the fit. It produces the same $\mathrm{RMSE} = 0.076$  as the standard SIS model and estimates an exponent $h$ close to one. Thus, the generalized model reduces to simple contagion, providing no evidence that a critical-mass threshold is required.

\begin{figure}
    \centering
    \includegraphics[width=0.6\linewidth]{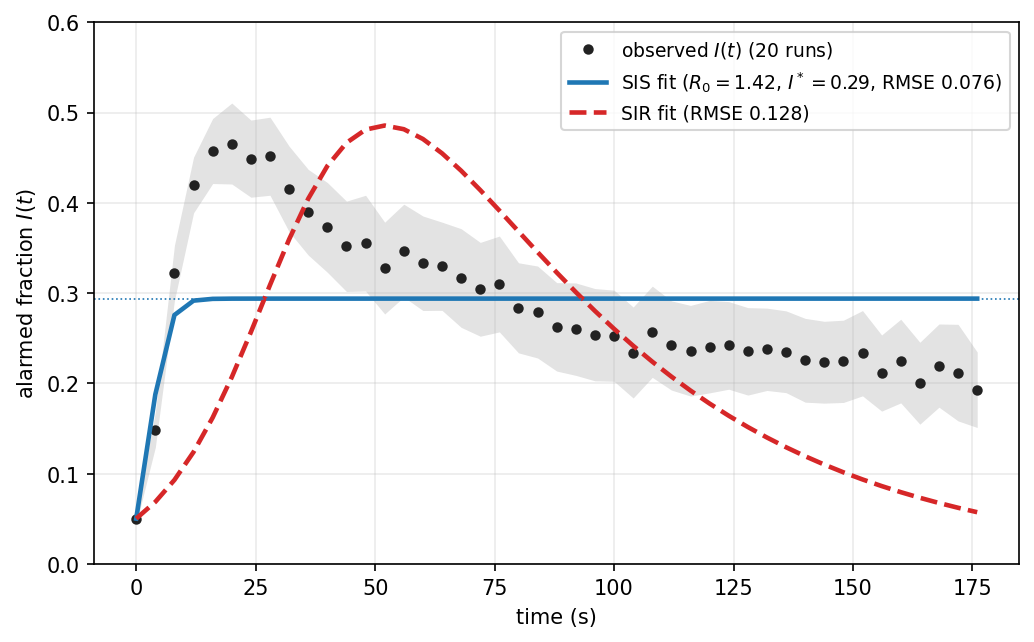}
    \caption{SIS and SIR fits to the alarmed fraction $I(t)$ across the Evacuation Shelter
runs. Each point is one 4\,s bin, giving the mean alarmed fraction across runs at that
moment; the shaded band is its standard error.}
    \label{fig:sir}
\end{figure}

H3 is partially supported. Neither the SIR nor the SIS model captures both phases of the observed trajectory using a single parameter pair, $(\beta,\gamma)$.  Figure~\ref{fig:sir} shows the fitted trajectories for both models.

\subsubsection{Ablation of Expression Modalities}~\label{sec:ablation}

\noindent\textbf{Design.} This experiment evaluates the contribution of each observable communication modality to affective contagion and tests \textbf{H4}. We conduct the ablation study in the Standing Line scenario, in which alarm is produced primarily through contagion. In contrast, the affective response in Evacuation and Concert is strongly influenced by the scenario context, and Contested Gate produces too little alarm for a reliable modality comparison.

In each ablation condition, we isolate one modality and mute the remaining modalities in both the agent prompts and the simulation. Agents continue to produce behaviors through all expression modalities, but other agents cannot perceive cues transmitted through the muted modalities. We evaluate facial expressions, gestures, vocalizations, and body motion. We exclude tactile communication from the analyses because tactile cues are controlled directly by the crowd simulator. Moreover, since agents do not collide in the Standing Line scenario,  tactile cues are rare or absent. Still, the tactile modality perception is muted in all conditions.

We compare four single-modality conditions against an all-cues baseline and an all-muted control.  All conditions use identical scenario parameters and are run 20 times each. The same 20 baseline crowds are evaluated under every condition to make sure the runs are consistent across replicates.

\minititle{Analysis.} For each condition and replicate, we compute the mean proportion of
non-seed agent appraisals in which the agent occupies the panic quadrant ($V<0$, $A>0.35$). We then average this measure across the 20 runs. We compare each
single-modality condition with the all-cues baseline to determine how much of the full
contagion effect it reproduces. We also compare each single-modality condition with the
all-muted control to determine whether access to that modality increases alarm above the
level observed when no expressive cues are perceptible. Because the same baseline crowd is
evaluated under every condition, both comparisons use paired $t$-tests on the
replicate-level differences across the 20 matched crowds.  We also summarize each modality as the fraction of baseline contagion retained: $(\mathrm{modality} - \mathrm{floor})/(\mathrm{baseline} - \mathrm{floor})$. Thus, the
all-cues baseline is scaled to $1$ and the all-muted floor to $0$.

Because voice is omnidirectional while faces and gestures are limited to the visual field,
part of a modality's total effect is imposed by the exposure geometry. Therefore, we
decompose each modality's alarmed fraction by exposure as:
\[
P(\text{panic}) = P(\text{exp})\,E[\text{panic} \mid \text{exp}]
                + P(\text{unexp})\,E[\text{panic} \mid \text{unexp}].
\]
Over all non-seed appraisal steps, pooled across the 20 runs of that modality's condition, $P(\text{exp})$ is the fraction of steps in which the agent perceives at least one neighbor displaying an alarm cue, and $E[\text{panic} \mid \text{exp}]$ is the fraction of those exposed steps in which the agent itself occupies the panic quadrant at the same step; $P(\text{unexp})$ and
$E[\text{panic} \mid \text{unexp}]$ are the corresponding quantities for steps with no
perceived cue.  However, this should be read as a descriptive accounting since $E[\text{panic} \mid \text{exp}]$ includes panic persisting from earlier exposures, and $P(\text{exp})$ reflects both the
modality's geometry and the amount of emission its own contagion level generates.

\minititle{Results.} With all communication modalities available, the mean panic-quadrant
fraction is $0.60$. In the all-muted condition, in which agents cannot perceive any
expressive cues from their neighbors, the mean fraction falls to $0.09$. Muting all
modalities therefore removes most of the observed contagion, indicating that alarm in this
scenario mainly depends on perceptible neighbor expressions. 

\begin{figure}
    \centering
    \includegraphics[width=0.5\linewidth]{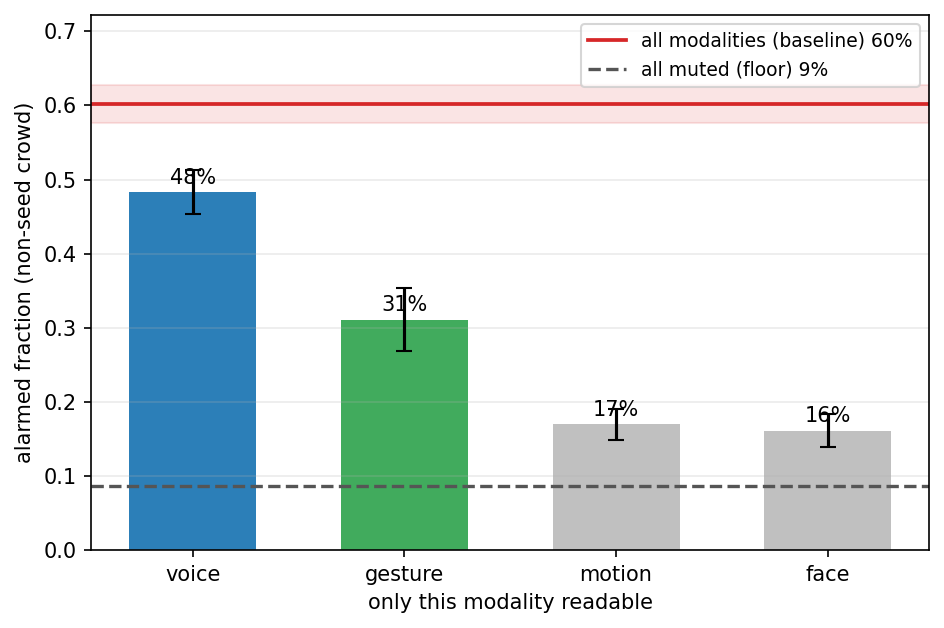}
    \caption{Perception modality ablation (Standing Line). Bars show the non-seed crowd's alarmed fraction when only one modality is readable, with the all-cues baseline and all-muted floor marked. Voice carries the most, followed by gesture, but no single modality reproduces the full spread.}
    \label{fig:ablation}
\end{figure}

No single modality reproduces the full all-cues effect. Voice produces the largest
panic-quadrant fraction at $0.48$, retaining $77\%$ of the floor-adjusted baseline effect.
Gesture produces a fraction of $0.31$, retaining approximately $43\%$. Motion and face
produce substantially smaller fractions of $0.17$ and $0.16$, retaining approximately
$16\%$ and $14\%$, respectively (Figure~\ref{fig:ablation}).

Each single-modality condition differs significantly from the all-muted control. The
paired-test results are $t=13.5$ for voice, $t=5.6$ for gesture, $t=4.4$ for motion, and
$t=5.0$ for face, with all $p<0.001$. Each single-modality condition also remains
significantly below the all-cues baseline. The corresponding results are $t=-4.0$ for
voice, $t=-5.7$ for gesture, $t=-23.2$ for motion, and $t=-17.2$ for face, with all
$p<0.001$. These results indicate that no single modality substitutes fully for the
combined cue set. 

Table~\ref{tab:channelDecomp} provides a descriptive account of modality reach and conditional panic. Voice has the greatest exposure rate, with alarm cues perceived in $53\%$ of appraisals. However, its conditional panic rate of $0.78$ is similar to gesture's rate of $0.80$. Therefore, voice  produces more total contagion simply because it is perceived more often.  Gesture and face provide a cleaner comparison because both share the same visual constraints. Gesture is perceived in more appraisals than face, $0.31$ versus $0.07$, and has a higher conditional panic rate, $0.80$ versus $0.60$. The agents' logged reasoning suggests that fearful faces are interpreted as another agent's private emotional state, such as ``looks fearful,'' whereas shielding or alarmed pointing is treated as evidence of an external threat, such as ``others shielding suggests a potential threat.'' 

In the table, $P(\text{exp}) = 0$ for motion. The alarm measure counts only fleeing on this
modality, and no agent perceives a fleeing neighbor. However, agents perceive
agitated motion, which the alarm measure does not count, and show agitation themselves on
$7.3\%$ of appraisals against $4.2\%$ in the all-muted control. The motion modality
therefore transmits, but through lower-intensity cues than the alarm measure captures.

\begin{table}[t]
  \centering
  \small
  \caption{modality effect decomposed by exposure, over the same $110$\,s window as the
  ablation. $P(\text{exp})$ is the fraction of non-seed appraisals perceiving an alarm cue
  through the modality; $E[\text{panic}\mid\text{exp}]$ and $E[\text{panic}\mid\text{unexp}]$
  are the panic-quadrant fractions among exposed and unexposed appraisals; Total is the
  overall alarmed fraction, equal to their exposure-weighted sum.}
  \label{tab:channelDecomp}
  \setlength{\tabcolsep}{6pt}
  \begin{tabular}{lcccc}
    \toprule
    Modality & $P(\text{exp})$ & $E[\text{panic}\mid\text{exp}]$
            & $E[\text{panic}\mid\text{unexp}]$ & Total \\
    \midrule
    Voice   & $0.53$ & $0.78$ & $0.15$ & $0.48$ \\
    Gesture & $0.31$ & $0.80$ & $0.08$ & $0.31$ \\
    Motion  & $0.00$ & ---    & $0.17$ & $0.17$ \\
    Face    & $0.07$ & $0.60$ & $0.13$ & $0.16$ \\
    \bottomrule
  \end{tabular}
\end{table}

H4 is supported as contagion is carried by observable modalities. When no modality is readable,
transmission collapses to a spontaneous floor, and each single modality lifts alarm
significantly above that floor while none reproduces the full spread alone. Voice's lead
in total contagion is consistent with its reach; per exposure, the gesture cues are
associated with the most panic.

\paragraph{Expression Modality Specialization}
\noindent\textbf{Design.} The ablation results show through which modalities agents catch
emotions. We also evaluate through which modalities agents most likely express their valence
and arousal. For this, we pool every appraisal across all Gemini experiments, excluding
the parameter-sweep variants, comprising $315{,}864$ time steps from $490$ runs.

\minititle{Analysis.} For each expression modality, we group the appraisal steps by the expression
category the agent selected on that modality and report the mean valence and arousal of
each category. To quantify the relationship between affect and expression choice, we report $\eta^2_V$ and $\eta^2_A$, the share of the variance in valence and arousal, respectively, that lies between expression categories. Finally, we count how many of the four modalities are simultaneously active at each step and relate this count to affect, testing whether stronger feelings activate more modalities at once. An active modality can be a non-neutral face, any gesture,
any vocalization, or a non-baseline movement.

\minititle{Results.} The expressions are mainly distributed along the high-arousal region
of the affective circumplex. Unpleasant, high-arousal expressions including screaming,
shouting, gasping, fearful or angry facial expressions, agitation, fleeing, and shielding,
cluster in the negative-valence corner. Pleasant, high-arousal expressions, such as
laughing and dancing, cluster in the positive-valence corner. Calmer and less intense
expressions including strolling, chattering, neutral facial expressions, and smiling are
closer to the lower-arousal portion of the circumplex.

The face emerges as the richest modality and the primary carrier of valence
($\eta^2_V = 0.77$; smile has mean $V = 0.62$, fearful $-0.80$ and angry $-0.47$, frown
and neutral in between). Face also tracks arousal ($\eta^2_A = 0.65$; fearful has mean
$A = 0.90$, angry $0.78$, and smiling $0.26$). Movement provides an equally clear arousal
axis ($\eta^2_A = 0.64$; in increasing order from strolling with mean $A = 0.11$,
purposeful $0.43$, agitated $0.83$, to fleeing $0.90$). The exception is idle, which
despite being the least active movement, has high arousal and negative valence
($A = 0.73$, $V = -0.62$). In the simulation logs, idle indicates a fear freeze instead of
calm stillness, consistent with freeze as a distinct, physiologically activated threat
response~\cite{roelofs2017freeze}.

Voice and gesture are sparsely used. However,  they occupy the extremes of the circumplex when used. Agents are silent on $57\%$ of steps and make no gesture on $69\%$, which dilutes their $\eta^2$ (voice $\eta^2_V = 0.50$, $\eta^2_A = 0.40$; gesture $\eta^2_V = 0.56$, $\eta^2_A = 0.49$). Scream has $V = -0.90$, $A = 0.96$ and laugh has $V = 0.87$, $A = 0.82$ for voice; shield has $V = -0.78$ and dance $V = 0.87$ for gesture. Voice and gesture therefore function as categorical signals of intense affect and are generally reserved for emotionally extreme states. These are also the two modalities that produce the strongest affective spread in the ablation experiment described in Section~\ref{sec:ablation}.

\begin{figure*}
    \centering
    \includegraphics[width=0.99\linewidth]{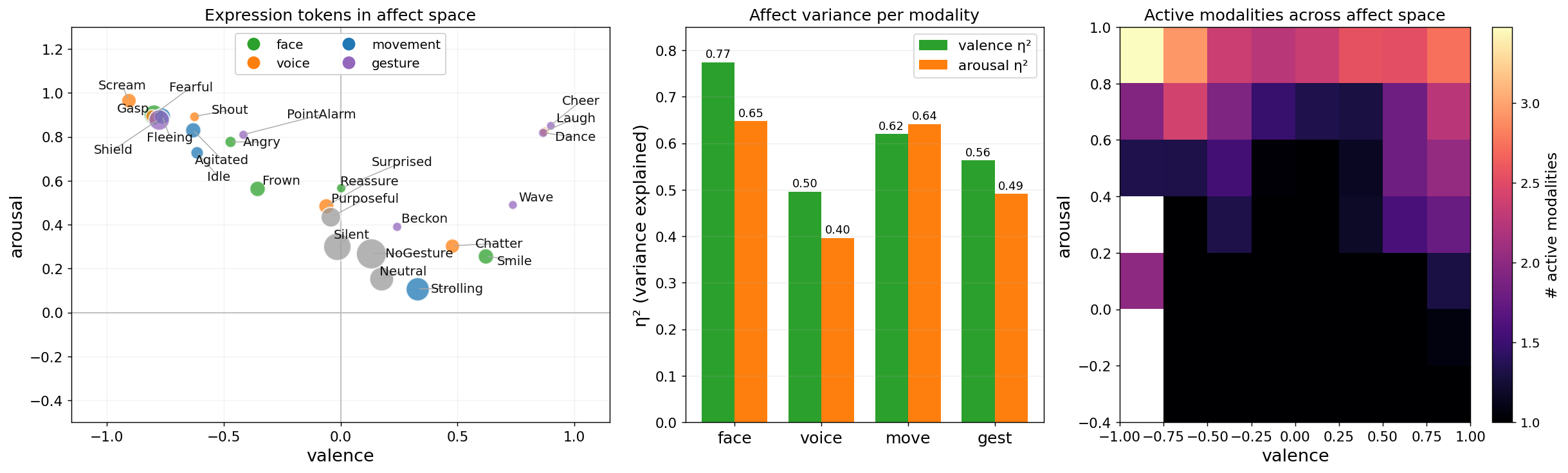}
    \caption{LLM-chosen expression across the valence--arousal space (pooled across all scenarios). \textbf{Left:} each expression token at the mean valence and arousal of the steps on which agents chose it, colored by modality (marker size $\propto$ frequency; gray = neutral defaults).\textbf{ Center:} variance in valence and arousal explained by each modality's token choice ($\eta^2$).\textbf{ Right:} mean number of active expression modalities across the affect space.}
    \label{fig:expressionChannels}
\end{figure*}

Expression becomes more multimodal as arousal increases. Splitting the appraisals into
low, mid, and high arousal bands (cut at $A = 0.33$ and $0.6$), the mean number of active
modalities rises from $0.5$ to $1.3$ to $2.9$ across the bands. Low-arousal states are
uncommon in the evaluated scenarios, with a median of $0.42$ arousal across all appraisals
and only $14\%$ of steps falling below zero arousal. This distribution reflects the design
of the scenarios, which involve excitement, panic, and interpersonal friction, but not
low-activation states such as boredom or sadness. Figure~\ref{fig:expressionChannels} summarizes the pooled results.

These associations are not driven by differences between scenarios. When $\eta^2$ is
recalculated separately within each scenario and the resulting values are averaged, the
estimates barely change: face $\eta^2_V = 0.72$, $\eta^2_A = 0.63$; movement $0.64$ and
$0.67$; voice $0.44$ and $0.38$; gesture $0.53$ and $0.47$. Thus, each expression modality
continues to track affect within individual scenarios as agents' own affective states
change. Within scenarios, movement is the strongest arousal readout.

\subsubsection{Emergent Behavior Analysis}
To understand how behaviors emerge based on personality and test \textbf{H5}, we run several scenarios.

\paragraph{Neurotic Crowd Under Ambiguous Alarm}~\label{sec:h5a}
\noindent\textbf{Design.} This experiment tests whether increased crowd neuroticism can turn a calm scene into a panic contagion under an ambiguous situation (\textbf{H5.a}). We use the Evacuation setup by eliminating panic seeds and thus removing any external cause for panic.  The crowd starts at neutral mood. We then vary only the crowd's mean neuroticism $0.3 / 0.6 / 0.9$. We run 20 iterations at each level with a different baseline crowd.

\minititle{Analysis.} For each run we measure the fraction of appraisals in the panic quadrant ($V < 0$, $A > 0.35$) and the fraction showing an external panic cue. We then test the trend of these values against neuroticism across the three ordered levels. To distinguish genuine contagion from agents merely being individually anxious, we also examine the time course and the lagged neighbor coupling as in~ Section~\ref{sec:localEmotionContagion}.

\minititle{Results.} At $N = 0.3$ the crowd largely treats the ambiguous alarm as routine,
with mean valence $V = -0.04$. $30\%$ of appraisals fall in the panic quadrant, but just past its boundary (mean $V = -0.25$ and $A = 0.54$ within the quadrant), indicating slight unease. Even so, the crowd remains composed without falling into collective panic. Only $4.0\%$ of appraisals show any overt alarm cue, distributed as $3.5\%$ fleeing, $2.0\%$ fearful faces, $1.2\%$ shielding or pointing in alarm, and $0.1\%$ screaming or gasping.

At $N = 0.6$ most of the crowd panics, with $83\%$ in the panic quadrant, $70\%$ showing
an alarm cue, and a mean valence $V = -0.48$. The cues are distributed as $63\%$
fearful faces, $62\%$ shielding or pointing, $61\%$ fleeing, and $27\%$ screaming or
gasping.

At $N = 0.9$ panic is near-total, with $92\%$ in the panic quadrant, $98\%$ showing an
alarm cue, and a mean valence $V = -0.71$. The cues are distributed as $97\%$ fearful
faces, $97\%$ shielding or pointing, $91\%$ fleeing, and $88\%$ screaming or gasping.

Every measure rises monotonically with neuroticism (Spearman $\rho = 0.94$;
Jonckheere--Terpstra trend test across the three ordered levels $z = 8.1$, $p < 0.001$, 20 runs per level, for both the panic-quadrant and alarm-cue fractions). Across all crowds, the alarmed fraction increases over time, with higher neuroticism producing both faster growth and higher saturation. The calm crowd climbs gradually, from $1\%$ in the first tenth of the run to $37\%$ near the end. The $N = 0.6$ crowd rises from $10\%$ to $97\%$ by mid-run, while the $N = 0.9$ crowd begins with $33\%$ alarmed and saturates at $100\%$ within the first half of the simulation.

\begin{figure*}
    \centering
    \includegraphics[width=0.99\linewidth]{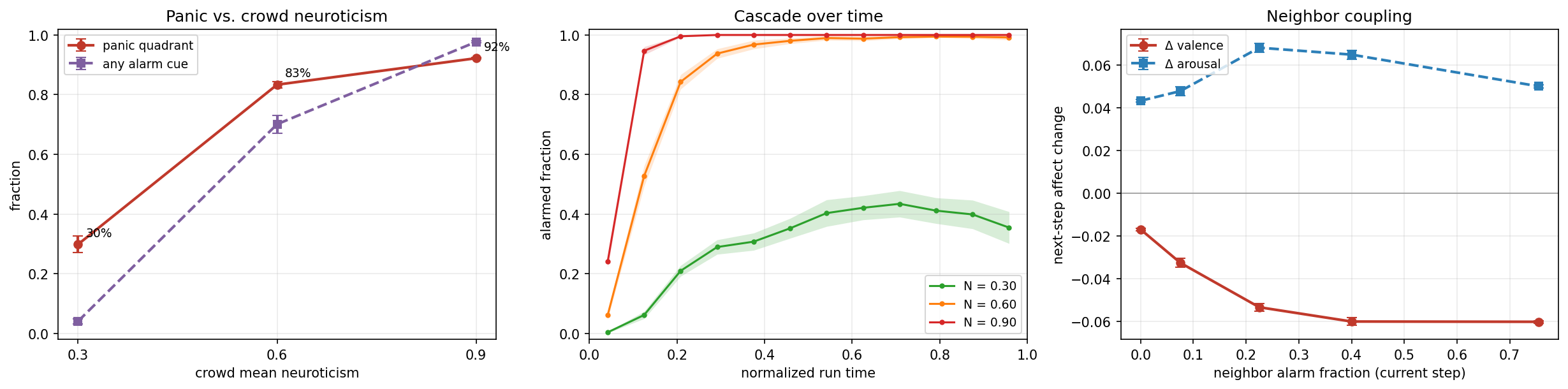}
\caption{Crowd neuroticism determines whether an ambiguous alarm ignites panic
(Evacuation). \textbf{Left:} fraction of appraisals in the panic quadrant and fraction
showing alarm cues, against mean crowd neuroticism. \textbf{Center:} the alarmed fraction
of appraisals over normalized run time. \textbf{Right:} the lagged neighbor coupling. As
the fraction of alarmed neighbors at the current step rises, an agent's next-step valence
falls and arousal rises.}
    \label{fig:neuroAlarm}
\end{figure*}

Consistent with contagion, an agent's affect at the next appraisal step is predicted by
its neighbors' current alarm ($\beta_1^V = -0.066$, $t = -7.2$, $p < 0.001$;
$\beta_1^A = 0.054$, $t = 6.6$, $p < 0.001$), controlling for the agent's own affective
state and local density. Figure~\ref{fig:neuroAlarm} summarizes the results. 

The gradual ignition of the alarm, its dependence on disposition, and the neighbor coupling are 
consistent with a contagion mechanism and support H5.a. Although the stable crowd is affected to some extent, with about a third of its appraisals in the panic quadrant, its response is bounded and distinct from the self-amplifying cascade that saturates the crowd at high neuroticism.

\paragraph{Calm Leaders}
\noindent \textbf{Design.} We hypothesize that including calm leaders in a crowd will
decrease crowd panic (\textbf{H5.b}). To test this, we run three configurations in
Evacuation with $k$ such agents ($k=0, 2, 4$) replacing ordinary members of a baseline
crowd of fixed size, seeded with a single panicked agent with $V = -0.85$, $A = 0.85$. Each leader's personality is drawn from a normal distribution centered on high conscientiousness, extraversion and agreeableness ($\mu_C,  \mu_E, \mu_{Agr} = 0.80$), and low neuroticism ($\mu_N = 0.15$), all with $\sigma = 0.04$, so that they tend to take charge, stay organized and helpful, and remain calm.  Although we do not explicitly designate these agents as leaders, we call them such because their disposition produces reassuring, beckoning, and purposeful behavior, which other agents can perceive and follow. The leaders start composed, with arousal $-0.4$ and slightly positive valence $0.1$, and perceive the same alarm as everyone else.   We take 20 runs per $k$, each with a different baseline personality draw.

\minititle{Analysis.} We calculate the non-leader crowd's mean arousal and valence, its
alarm-cue fraction across all modalities, and its panic-quadrant fraction, with the seed and
the leaders excluded. We test the trend of each measure against the number of leaders
across the three ordered levels.

\begin{figure}[t]
  \centering
  \includegraphics[width=0.6\linewidth]{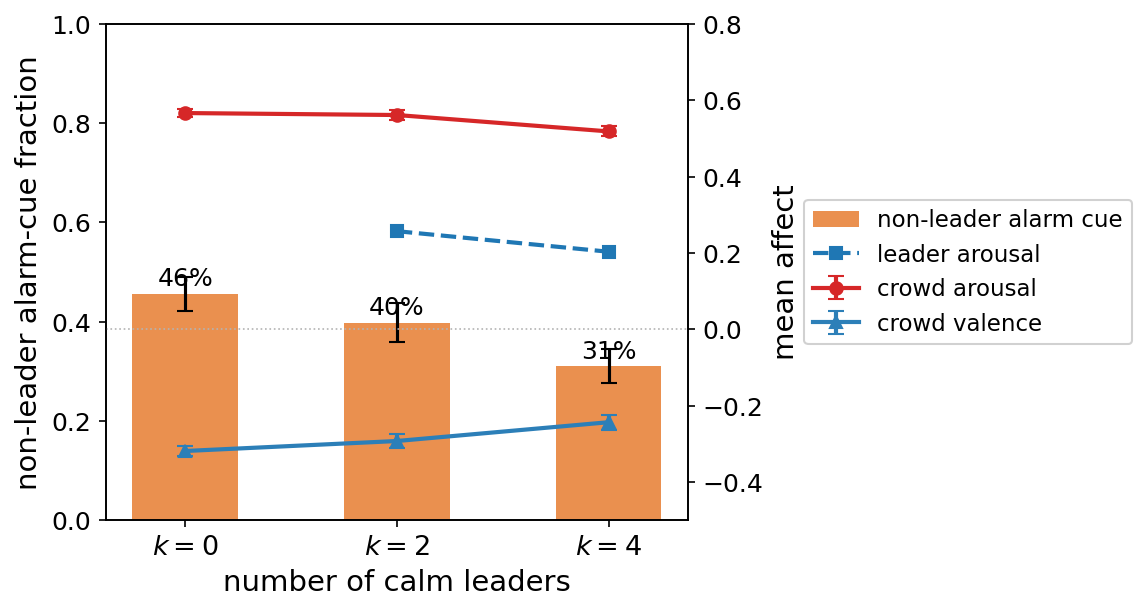}
  \caption{Calm leaders stay composed and modestly dampen the surrounding crowd's panic (Evacuation). Non-leader alarm cue fraction, crowd mean arousal, valence, and calm leader arousal based on number of calm leaders. }
  \label{fig:calm_leaders}
\end{figure}

\minititle{Results.} The seed reliably creates baseline panic. At $k=0$ the crowd has mean $A= 0.57$ and $V = -0.32$, with $46\%$ of appraisals showing an overt alarm cue and $77\%$ being in the panic quadrant of affect. The leaders hold their composure, with mean arousal of $0.26$ at $k=2$ and $0.20$ at
$k=4$, far below the surrounding crowd's $0.52$--$0.57$.  They intervene actively, calling out reassurance in $98\%$ of leader appraisals at $k=2$ and $100\%$ at $k=4$. Their calm measurably spreads. Excluding the leaders themselves, the non-leader alarm-cue fraction falls from $46\%$ to $40\%$ to $31\%$ as $k$ rises (Spearman $\rho=-0.36$; Jonckheere--Terpstra trend test $z=2.8$, $p=0.005$, 20 runs per level). The panic-quadrant fraction falls from $77\%$ to $75\%$ to $69\%$ ($\rho=-0.40$; $z=3.1$,  $p=0.004$). The underlying mean arousal decreases from $0.57$ to $0.52$, and mean valence increases from $-0.32$ to $-0.25$. The effect is modest, as four calm leaders still leave $69\%$ of the crowd in panic. Figure~\ref{fig:calm_leaders} depicts the results. H5.b is supported, as calm leaders modestly reduce, though do not prevent, the surrounding crowd's panic. 

\paragraph{Disagreeable Crowd with Angry Provocateurs}
\textbf{Design.} We ask whether agents seeded with angry behaviors tip a crowd toward anger (\textbf{H5.c}). For this, we use the Contested Gate scenario, where the agents are initialized with the goal to hurry to the other side through a narrow gate. We add provocateurs at levels $k = 0, 2, 4$. The crowd is composed of 18 agents in every condition, where the $k$ provocateurs replace ordinary agents. The provocateurs have low agreeableness ($\mu_{Agr} = 0.12$), raised neuroticism ($\mu_N = 0.60$), raised extraversion ($\mu_E = 0.70$), and lowered conscientiousness
($\mu_C = 0.35$), all with $\sigma = 0.05$, and are seeded into the unpleasant-activated
region ($V = -0.6$, $A = 0.45$). We run the same situation with a baseline crowd (all trait means 0.5, $\sigma = 0.16$) and a disagreeable crowd ($\mu_{Agr} = 0.30$, $\mu_N = 0.35$, other trait means 0.5; $\sigma = 0.16$). The disagreeable crowd is deliberately kept at moderately low neuroticism as we are interested in observing provocation as anger instead of anxiety. We take 20 runs for each provocateur level. 

\minititle{Analysis.} For each run, we measure the crowd's anger and fear cue
fractions, excluding the seeded provocateurs. Since anger and fear occupy the same
affective quadrant, instead of directly measuring the affective state we look at outward
expressions to better understand how agents appraise the situation. Anger cues include
angry faces and shouts; fear cues include fearful faces, screaming or gasping, fleeing,
and shielding or pointing in alarm. We test the trend of each cue fraction against the
number of provocateurs $k$ across the three ordered levels. The same setup is run for both
crowd compositions, and we test whether the crowds respond differently by regressing the
cue fraction on crowd type, $k$, and their interaction.

\begin{figure}
    \centering
    \includegraphics[width=0.5\columnwidth]{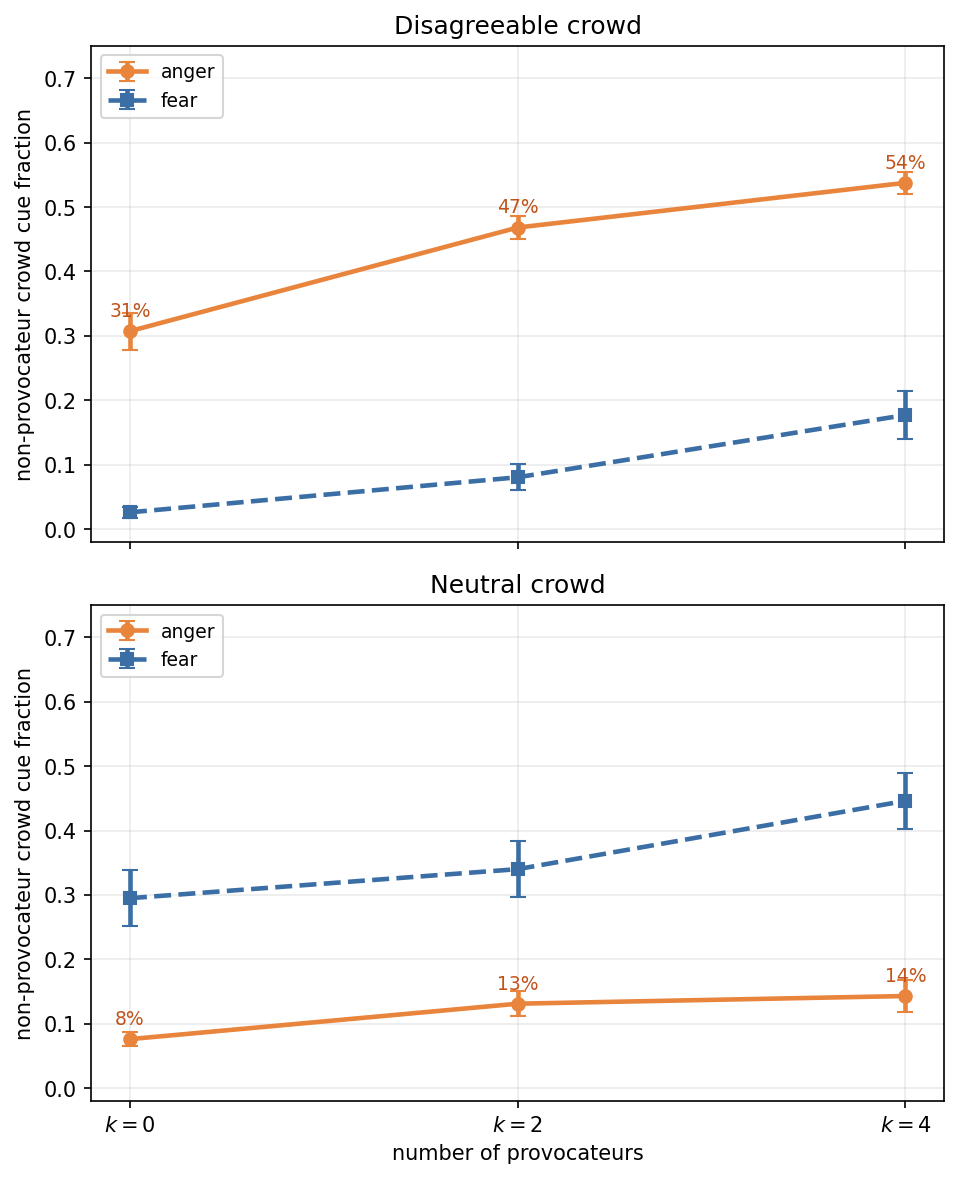}
\caption{Anger contagion is conditional on the receiver's temperament (Contested Gate).
Fraction of non-provocateur appraisals showing fear cues (blue) and anger cues (orange),
against the number of provocateurs. \textbf{Top:} disagreeable crowd. \textbf{Bottom:}
baseline crowd.}
    \label{fig:angerContagion}
\end{figure}

\minititle{Results.} The disagreeable crowd's anger rises with the number of
provocateurs, from $31\%$ to $47\%$ to $54\%$ (Spearman $\rho = 0.72$;
Jonckheere--Terpstra trend test across the three ordered levels $z = 5.8$, $p < 0.001$,
20 runs per level). Fear also increases with the provocateurs but stays mild, changing
from $3\%$ to $8\%$ to $18\%$ ($\rho = 0.50$; $z = 3.9$, $p < 0.001$). The baseline crowd
catches far less anger, rising only from $8\%$ to $13\%$ to $14\%$. This increase is weak
but significant ($\rho = 0.27$; $z = 2.1$, $p = 0.039$), well below the disagreeable
crowd's. For the agents in the baseline crowd, fear is the dominant emotion, changing from
$30\%$ to $34\%$ to $45\%$, well above their anger, and also rising with the number of
provocateurs ($\rho = 0.30$; $z = 2.3$, $p = 0.031$). According to the logs, these agents
interpret the hurried, congested gate as threatening.
A crowd-type $\times$ provocateur-count interaction tests whether the two crowds respond
differently. In the disagreeable crowd, the anger-cue fraction increases by $0.054$ per
added provocateur; in the baseline crowd, by only $0.017$. The difference between these
slopes is the statistically significant interaction effect ($\beta = 0.037$, $t = 3.3$,
$p = 0.004$). For fear, the slopes are indistinguishable ($p = 0.91$). Disposition is
therefore associated specifically with a stronger anger response, not with heightened
reactivity in general.
The results show that, since the two emotions share the same affective quadrant, it is the
appraiser's personality that settles which one the crowd catches. Contagion of anger
expression requires a blameable source and anger-prone receivers. Therefore, H5.c is
supported. Figure~\ref{fig:angerContagion} displays the fear and anger spread within
disagreeable and baseline crowds.

\subsection{Appraisal Validation}
\label{sec:appraisal}
 Since every affective change in the model is produced by a stochastic language model, we
characterize the appraisal step in isolation and verify that it is systematic and bounded.
The probe runs outside the spatial simulation, separating the cue's effect from the crowd
dynamics, and measures the input--output behavior of a single appraisal. It uses the
simulation's own prompt and parser and varies one input at a time. Each scene places the
agent in a small local group (5 others in the calm baseline, 6 in the cue scenes),
matching the neighborhood an agent typically perceives in the simulation.

The probe covers the two halves of the perception--appraisal--affect--expression loop. In
the appraisal half, the agent's own affect is held neutral and the perceived cue varies
across a calm baseline and an alarm scene, a joyous scene, and an angry scene, each with others
showing the corresponding cues. In the generation half, the perceived scene is held benign
and the agent's affect is set to panicked, elated, calm, or angry, and we read the
distribution over chosen expression categories. Every condition is run for a
low-neuroticism ($0.15$) and a high-neuroticism ($0.85$) profile, with all other traits at
$0.5$. Each direction $\times$ input $\times$ personality cell is sampled $N = 40$ times
at the operating temperature of $0.4$. The temperature is additionally swept over $0.0$,
$0.4$, and $0.8$ to assess reliability.

We further test robustness to prompt construction by repeating the probe under seven
prompt variants. These comprise the base prompt, two semantic paraphrases (plain and
formal) that leave the JSON contract unchanged, and four ablations that drop the
personality block, the situational context, the outward-behavior clause, or the
social-framing line. The full prompts are provided in Appendix~\ref{app:promptAppraisal}.
Finally, we evaluate the robustness of appraisal across models.

\subsubsection{Appraisal and Generation Probes}
 The same cue is appraised differently according to disposition. Averaging over the $N=40$
samples per cell, the identical alarm scene drives the neurotic agent to $V=-0.60$,
$A=0.80$ and the stable agent only to $V=-0.20$, $A=0.37$. The angry scene shows the same
gap, with $V = -0.40$, $A = 0.60$ for the neurotic agent and $V = -0.10$, $A = 0.20$ for
the stable one. The calm baseline leaves the neurotic agent slightly uneasy ($V = -0.13$,
$A = 0.25$) and the stable agent mildly content ($V = 0.08$, $A = 0.00$). Disposition even
reverses the sign of joy. The stable agent appraises the celebratory scene as pleasant
($V = 0.20$, $A = 0.10$) and the neurotic one as mildly negative and more activated
($V = -0.18$, $A = 0.38$). The reasoning logs show the neurotic agent reading the
high-energy scene as faintly threatening. Table~\ref{tab:appraisalDir} summarizes the
results.

In the generation direction, disposition again separates the agents. Given an identical
panicked internal state, the neurotic agent shows a distressed face and agitated movement
on every sample. The stable agent keeps a neutral face and moves purposefully, masking its
internal state. Disposition also determines gesture, with the neurotic agent shielding
itself when panicked or angry and the stable agent making none. Voice stays silent
throughout the generation probe. In the simulation, vocalization accompanies a provoking
scene, such as an alarm sounding or others screaming. The probe holds the perceived scene
benign, and no internal state alone, however extreme, produces a vocalization. The perceived context thus controls how much of the felt state is displayed, and full-intensity expressions require the scene to warrant them. Table~\ref{tab:genDir} summarizes the results.

\begin{table}[t]
  \centering
  \caption{Appraisal direction (perception $\rightarrow$ affect): mean $\pm$ std of
  the returned valence and arousal ($N=40$ per cell, at $T = 0.4$).}
  \label{tab:appraisalDir}
\small
  \begin{tabular}{llcc}
    \hline
    Crowd Cues & Personality & Valence & Arousal \\
    \hline
    Calm baseline        & stable   & $0.08 \pm 0.02$ & $0.00 \pm 0.02$ \\
    Calm baseline        & neurotic & $-0.13 \pm 0.05$ & $0.25 \pm 0.05$ \\
    Alarm  & stable   & $-0.20 \pm 0.00$ & $0.37 \pm 0.05$ \\
    Alarm  & neurotic & $-0.60 \pm 0.00$ & $0.80 \pm 0.00$ \\
    Joy    & stable   & $0.20 \pm 0.02$ & $0.10 \pm 0.00$ \\
    Joy    & neurotic & $-0.18 \pm 0.07$ & $0.38 \pm 0.05$ \\
    Anger       & stable   & $-0.10 \pm 0.00$ & $0.20 \pm 0.00$ \\
    Anger       & neurotic & $-0.40 \pm 0.00$ & $0.60 \pm 0.00$ \\
    \hline
  \end{tabular}
\end{table}

\begin{table}[t]
  \centering
\small
  \caption{Generation direction (affect $\rightarrow$ expression): dominant expression
  per modality chosen from an identical internal state, by personality ($N=40$ per
  cell at $T=0.4$). Voice is silent in all cases. Panicked internal state has $V{=}{-}0.6$, $A{=}0.7$; elated $V{=}0.6$, $A{=}0.6$, calm $V{=}0.2$, $A{=}{-}0.4$, and angry $V{=}{-}0.5$, $A{=}0.5$.}
  \label{tab:genDir}
  \setlength{\tabcolsep}{3pt}
  \begin{tabular}{lllll}
    \hline
    Internal state & Personality & Face & Movement & Gesture \\
    \hline
    Panicked  & stable   & Neutral (100\%) & Purposeful (100\%) & None (100\%) \\
    Panicked  & neurotic & Frown (100\%)   & Agitated (100\%)   & Shield (100\%) \\
    Elated       & stable   & Neutral (100\%) & Strolling (92\%)   & None (100\%) \\
    Elated       & neurotic & Neutral (100\%) & Purposeful (80\%)  & None (100\%) \\
    Calm      & stable   & Neutral (100\%) & Strolling (100\%)  & None (100\%) \\
    Calm      & neurotic & Neutral (100\%) & Strolling (88\%)   & None (100\%) \\
    Angry    & stable   & Neutral (100\%) & Purposeful (100\%) & None (100\%) \\
    Angry   & neurotic & Frown (100\%)   & Agitated (100\%)   & Shield (100\%) \\
    \hline
  \end{tabular}
\end{table}
 




\subsubsection{Temperature Robustness} 
Sweeping the sampling temperature confirms that the appraisal direction is robust to
temperature. The mean within-condition standard deviation for valence is $0.00$ at $T=0$,
$0.025$ at the operating $T=0.4$, and $0.028$ at $T=0.8$. For arousal, it is $0.00$ at
$T=0$, $0.02$ at the operating $T=0.4$, and $0.036$ at $T=0.8$. The generation direction
is equally stable. Within each cell, we measure agreement as the share of the 40 samples choosing that cell's most frequent expression category. Agreement is $1.00$ for face, gesture, and
voice at every temperature. For movement, the agreement falls only from $1.00$ at $T=0$ to $0.95$ at $0.4$ and $0.93$ at $0.8$.  The model commits to nearly the same appraisal target and the same expression choices across temperatures, suggesting that the emergent dynamics hold without a finely tuned sampling temperature.

\subsubsection{Prompt Robustness} 
\begin{figure*}[t]
  \centering
  \includegraphics[width=\textwidth]{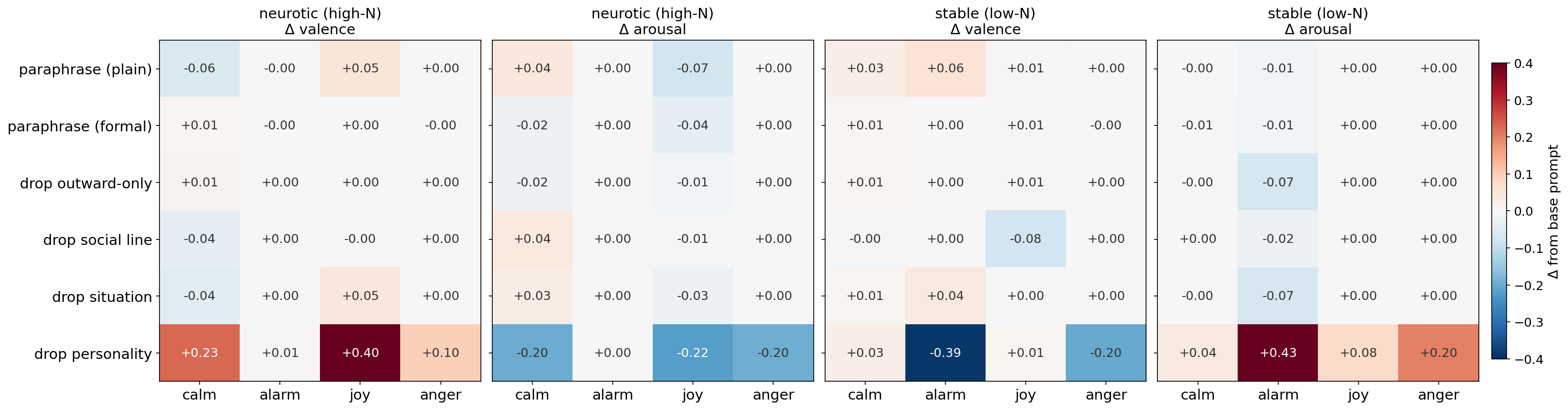}
  \caption{Prompt robustness for \texttt{gemini-3.1-flash-lite} at $T=0.4$. Change in returned
valence  and arousal  from the base prompt, for
every prompt variant and cue, for the neurotic (high-N, \textbf{left two}) and stable
(low-N, \textbf{right two}) dispositions. The paraphrases are meaning-preserving
rewordings of the base instructions, one in plain language and one more formal.
The ablations each drop a single prompt component (Appendix~\ref{app:promptAppraisal}). }
  \label{fig:probePrompts}
\end{figure*}

The results suggest that plain versus formal wording makes essentially no difference. Both
paraphrases place all four cues within $0.06$ valence of the base prompt for both
dispositions and preserve the disposition gap. We calculate $\Delta V$ and $\Delta A$ as
the mean neurotic-minus-stable difference in returned valence and arousal. The base prompt
yields $\Delta V = -0.20$, $-0.40$, $-0.40$, and $-0.30$ and $\Delta A = 0.24$, $0.43$,
$0.30$, and $0.40$ for the calm, alarm, joy, and anger cues, respectively. Removing the
personality block collapses every gap to $|\Delta V|, |\Delta A| < 0.01$, confirming that
the trait manipulation acts through the prompt. The personality ablation also clarifies
the contribution of disposition. Even with no personality supplied, the unambiguous alarm
cue drives a strong negative response, so the cue itself sets the direction and
disposition scales the intensity. For ambiguous positive cues, disposition is decisive, as
neuroticism reverses the sign of calm and joy. The remaining three ablations have little
effect. Dropping the outward-only clause, the social-framing line, or the situational
context barely moves any cue. Under these probe conditions, appraisal changes little when these framing elements are removed. Figure~\ref{fig:probePrompts} summarizes the comparison, plotting
each variant's shift from the base prompt.
 
\subsubsection{Robustness Across Models}
\label{sec:models}
All the tests so far use \texttt{gemini-3.1-flash-lite}. Since the architecture is
backend-agnostic, we ask whether the appraisal behavior survives a change of model. We run
the appraisal probe against four backends. These are \texttt{gemini-3.1-flash-lite}, a
second proprietary model (\texttt{gpt-4o-mini}\footnote{OpenAI Chat Completions,
\texttt{json\_object} output; temperature $0.4$, $200$ tokens; up to six concurrent calls,
retried on HTTP $429$/$503$/$529$ with back-off.}), a mid-size local model
(\texttt{llama3.1:8b}\footnote{Local Ollama (\texttt{/api/chat}); JSON requested in the
prompt; temperature $0.4$, $200$ max tokens; single attempt per call.}), and a small one
(\texttt{llama3.2:3b}\footnote{Local Ollama (\texttt{/api/chat}), configured similarly to
\texttt{llama3.1:8b}.}), each swept over sampling temperatures $\{0, 0.4, 0.8\}$ with
$N = 40$ samples per cell. We compare the backends on four criteria. Format reliability is
the fraction of generations that parse into the schema. Sampling reliability is the
within-condition standard deviation of the returned valence and arousal. Personality
dependence is the same-cue gap in returned affect between the two dispositions.
Directional validity marks whether the returned affect places each cue in its expected
region of the circumplex, with joy more pleasant than the calm baseline and alarm and
anger more unpleasant and more aroused.

\begin{figure*}
  \centering
  \includegraphics[width=\linewidth,
      trim={0cm 0cm 0cm 1cm},
    clip
  ]{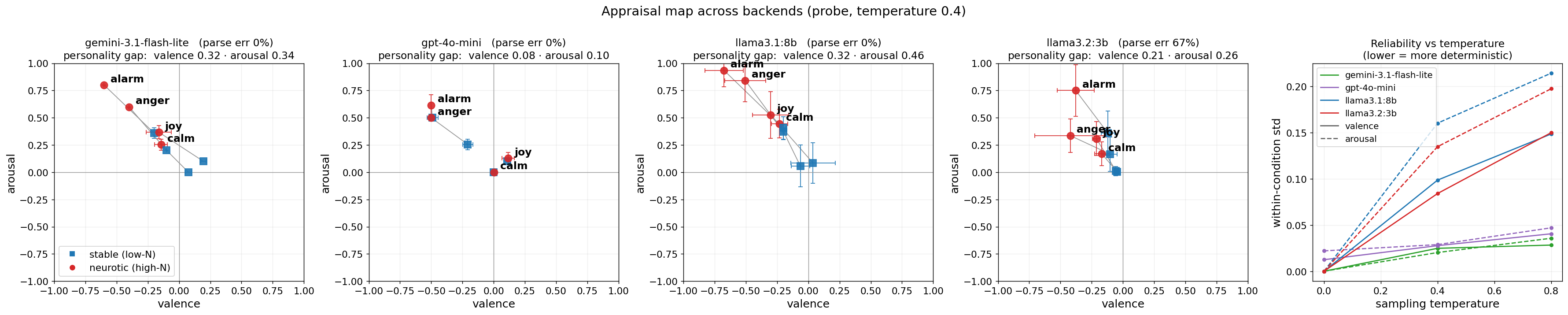}
  \caption{Appraisal map across four LLM backends. \textbf{First four panels:} returned valence/arousal for the four cues at sampling temperature 0.4, for the stable (squares) and neurotic (circles) personalities, each panel annotated with its parse-error rate and mean stable-versus-neurotic gap. \textbf{Rightmost panel:} within-condition standard deviation of the returned valence (solid) and arousal (dashed) against sampling temperature. The hosted models, \texttt{gemini-3.1-flash-lite} (green) and \texttt{gpt-4o-mini} (purple), stay near-deterministic on both axes at every temperature, whereas both local models' variance rises (blue and red). }
\label{fig:models}
\end{figure*}

Reliability and directional validity hold for the three capable backends and fail only at
the smallest. The personality dependence that controls the emergent dynamics, however,
varies sharply among them (Table~\ref{tab:models}, Fig.~\ref{fig:models}).
\texttt{llama3.1:8b} matches Gemini qualitatively. It produces no parse failures, returns
the correct cue ordering, and shows a personality gap of $|\Delta V| = 0.32$,
$|\Delta A| = 0.46$ against Gemini's $|\Delta V| = 0.32$, $|\Delta A| = 0.34$, with every
per-cue gap significant for both backends (all Holm-adjusted $p < 0.001$). It differs
mainly in noise, with $3$--$8\times$ higher standard deviation. \texttt{gpt-4o-mini} is as
clean as Gemini, with no parse failures and near-deterministic sampling
($\sigma_V = \sigma_A = 0.03$). It orders the cues correctly and reads the alarm scene as
strongly unpleasant and activated ($V = -0.49$, $A = 0.56$). However, it nearly erases the
personality gap on the calm, alarm, and joyous scenes ($|\Delta V| \le 0.01$,
Holm-adjusted $p \ge 0.64$), retaining disposition sensitivity only on the anger cue
($\Delta V = -0.29$, $\Delta A = 0.24$, $p < 0.001$) and in a small arousal gap under
alarm ($\Delta A = 0.11$).

The smallest model, \texttt{llama3.2:3b}, is not a usable backend. $67\%$ of its
generations fail to produce schema-valid JSON, and the ones that parse order the cues
incorrectly. For instance, it interprets the joyful scene as mildly negative
($V = -0.06$). On the $33\%$ of generations that parse, it nonetheless returns a
personality gap ($|\Delta V| = 0.21$, $|\Delta A| = 0.26$) in the same direction as the
capable backends, significant on six of eight cue--dimension tests after correction
($p \le 0.011$) and marginal on the anger cue ($p = 0.051$). The estimate is
selection-biased, as parse failures are strongly disposition-dependent, with the
stable-disposition prompts parsing on $68$--$100\%$ of samples per cue and the
neurotic-disposition prompts on only $18$--$30\%$.

Gemini remains the best calibrated for our purposes as the only hosted backend strong on
both reliability and disposition sensitivity, and also the fastest. Simulations using \texttt{gemini-3.1-flash-lite} and \texttt{gpt-4o-mini} run in real time, with per-call appraisal latencies of approximately 0.75~s and 1.5~s, respectively, over the API. When run locally on an Apple M4 with 32~GB of memory, \texttt{llama3.2:3b} and \texttt{llama3.1:8b} have per-call appraisal latencies of approximately 1.1~s and 2.0~s, respectively.

\begin{table}[t]
  \centering
  \small
  \caption{Appraisal probe across four backends ($N = 40$ per cell; parse errors over all temperatures, other columns at temperature $0.4$). \emph{Personality gap} is the mean per-cue absolute difference $|\text{neurotic} -\text{stable}|$ in returned valence ($|\Delta V|$) and arousal ($|\Delta A|$), averaged
over the four cues. \emph{Dir.} (directional validity) marks whether each cue lands in its expected region of the circumplex relative to the calm baseline. E.g. joy should be more pleasant, and alarm and anger should both be more unpleasant and more aroused.}
  \label{tab:models}
  \setlength{\tabcolsep}{2pt}
  \begin{tabular}{lccccccc}
    \toprule
    Backend & Params & Parse err. & $\sigma_V$ & $\sigma_A$ & Gap $|\Delta V|$ & Gap $|\Delta A|$ & Dir. \\
    \midrule
    \texttt{gemini-3.1-flash-lite} & --- & 0\%  & 0.03 & 0.02 & 0.32 & 0.34 & \checkmark \\
    \texttt{gpt-4o-mini}           & --- & 0\%  & 0.03 & 0.03 & 0.08 & 0.10 & \checkmark \\
    \texttt{llama3.1:8b}           & 8\,B & 0\%  & 0.10 & 0.16 & 0.32 & 0.46 & \checkmark \\
    \texttt{llama3.2:3b}           & 3\,B & 67\% & 0.08\textsuperscript{*} & 0.14\textsuperscript{*} & 0.21\textsuperscript{*} & 0.26\textsuperscript{*} & \ding{55} \\
    \bottomrule
  \end{tabular}
  \\[2pt]
  {\footnotesize \textsuperscript{*}Computed on the 33\% of generations that parsed; the remainder failed the schema.}
\end{table}

\paragraph{Model Dependence of the Emergent Wave}
The probe shows that the appraisal map transfers reasonably well across backends. However,
since \texttt{gpt-4o-mini} exhibits little sensitivity to personality, we also test its
effect within the crowd simulation. To this end, we run the Standing Line wave using
\texttt{gpt-4o-mini} across 20 runs while holding the scenario, geometry, and initial
seeds constant. Despite sufficient throughput, the wave fails to emerge. No non-seeded
agent enters the panic quadrant in any of the twenty \texttt{gpt-4o-mini} runs, whereas
every one of the corresponding Gemini runs develops a wave, with on average $23\%$ of non-seed
appraisals falling in the panic quadrant. The seed cluster attenuates its
injected panic to a mean valence $V = -0.31$, compared with $V = -0.94$ under the Gemini
backend, and expresses essentially no outward alarm ($3\%$ of appraisals, compared with
Gemini's $98\%$). The backend's flattening of personality thus acts at the source, not letting the
seed cluster's high neuroticism sustain the panic expression that would ignite the wave. Because
the crowd is never exposed to an alarm cue, these runs establish the source failure and
leave receiver-side susceptibility untested. The local backends cannot be evaluated under
comparable conditions. \texttt{llama3.2:3b} is excluded by its parse failures, and a local
\texttt{llama3.1:8b} instance serializes concurrent requests, completing only about one
appraisal per agent per run, too few appraisal cycles to sustain meaningful contagion.

\section{Discussion}

Across the experiments, we observe contagion with spatial, temporal, and dispositional structure, produced entirely by per-agent appraisal.  Classical contagion theories  treat affect transfer as a form of automatic mimicry, in which an observed expression triggers a matching internal state directly and largely independent of who is doing the observing~\cite{hatfield1994emotional}. An alternative tradition, the social identity model of crowd behavior, argues instead that collective affect and action are governed by a shared group identity and situational norms~\cite{reicher1996battle, drury2018role}.  Our architecture contains neither mechanism, as there is no mimicry loop that copies a perceived expression, and agents carry no representation of group identity or shared norms. Instead, every transfer is mediated by an appraisal step that interprets a neighbor's expression in light of the agent's own personality, memory, and situational context. This approach is more in line with appraisal theory~\cite{scherer2001appraisal, lazarus1991progress}, especially social appraisal accounts of interpersonal emotion transfer~\cite{parkinson2011interpersonal, manstead2001social},  which posit that others' emotions can inform our interpretation of events.

We acknowledge that some degree of coupling is inherent by construction. The appraisal prompt exposes each agent to its neighbors' observable behavior, so the existence of emotional contagion is not itself the finding. What the architecture does not prescribe is the structure of that contagion, including how it propagates through space, where it stabilizes, which expression modalities dominate transmission, or whether contagion emerges at all under a given crowd composition. The modality-ablation experiment makes this explicit: when all communication modalities are disabled, alarm collapses to an approximately 9\% spontaneous baseline, demonstrating that transmission depends on observable expressions rather than arising as an artifact of the prompt itself.

We now turn to the individual hypotheses, asking in each case what the appraisal-mediated account explains and where it falls short.

H1 proposes that observable expressions from neighboring agents predict changes in an agent's affect. The results support this hypothesis for negative contagion and partially for positive contagion, and the strength of the evidence differs across scenarios. In Evacuation, greater exposure to alarm cues is associated with lower valence and higher arousal in the crowd-level trajectory, the run-level cross-sectional comparison, and the pooled lagged regression. However, because the runs are short, the lagged model cannot clearly distinguish step-to-step contagion from differences between runs or general changes over time. The evidence for H1.a therefore rests primarily on the trajectory and cross-sectional results. In Concert, valence and arousal increase over the course of each run, and agents in positive neighborhoods show higher valence and arousal than those in neutral neighborhoods. The lagged regression also shows that neighbor positivity predicts an increase in next-step valence. However, the corresponding arousal coefficient is small and negative. Thus, the increase in crowd-level arousal is likely to be due to the celebratory setting instead of a lagged transfer of arousal between neighbors. The lagged analysis therefore supports H1.b for valence but not for arousal. The Calm Plaza control shows no crowd-wide contagion because alarm cues are too infrequent to spread through the crowd. Nevertheless, on the relatively rare occasions when an agent perceives an alarmed neighbor, neighbor alarm predicts lower next-step valence and higher next-step arousal. Unlike in Evacuation, these associations remain stable after controlling for run fixed effects and within-run time trends. The affective-coupling mechanism therefore remains active, but its influence at the crowd level depends on whether sufficiently frequent and consistent cues are available to propagate.

H2, which hypothesizes that alarm from a localized seed spreads as a
traveling wave, is supported in the Standing Line scenario. Alarm onset
rises with distance from the seed cluster within runs, the fitted wavefront slope
excludes zero with a propagation speed of ${\approx}1$\,m/s, and the
steady-state gradient falls off significantly with distance. The front and
its outward direction emerge because each agent appraises only its
immediate neighbors and the elongated geometry makes that coupling local.
Traveling and stop-and-go waves are a familiar consequence of local
interaction in pedestrian dynamics~\cite{helbing2000escape}, and qualitatively similar fronts have
been observed in real crowd disasters~\cite{helbing2007dynamics}. Similarly, escape cascades sweep through fish schools from a single spontaneously startling individual, propagating
along the local visual network with a finite speed~\cite{rosenthal2015revealing}.
The stadium Mexican wave is a human analogue, traveling through a crowd as
a finite-speed front~\cite{farkas2002mexican}.

H3, which proposes that the crowd's alarmed fraction follows an epidemic-style contagion
model, is partially supported. Because agents can return to a calm state and be alarmed
again, the alarmed fraction settles at a nonzero endemic plateau instead of decaying to
zero, a pattern qualitatively consistent with an SIS process. The SIS model fits the
observed trajectory more closely than the SIR model, in which the infected fraction must return to zero. However, no single pair of transmission and recovery parameters captures both the early
overshoot and the subsequent endemic plateau. Moreover, recovery in our model emerges from
agents' repeated appraisals of their surroundings, not from a fixed decay rate. Extending
the transmission function to include a threshold effect does not improve
the fit, either. The best-fitting exponent remains close to one, so the generalized model reduces
to standard simple contagion. We therefore find no evidence that a critical mass is
required for alarm to spread under these conditions, and the threshold mechanism used in
some personality-based crowd models~\cite{Durupinar2016-va,dodds2005generalized} stays
idle in this setting. LLM-driven agents have previously reproduced epidemic-like population dynamics~\cite{williams2023epidemic}, although that work uses an explicitly specified infection process rather than an emergent affective-contagion mechanism.

H4, which states that contagion is carried by specific observable modalities, is supported.
The strongest evidence is that muting all modalities collapses the contagion to a near-spontaneous floor. Additionally, the modalities are complementary, each increasing the contagion significantly on its own, without a single modality reproducing the full contagion. Differences among modalities reflect their perceptual reach and how agents interpret their meaning. Voice produces the greatest overall spread because it is omnidirectional and reaches more agents. Gesture and facial expression share the same visual constraints, yet gesture produces substantially more contagion. The agents' logged reasoning suggests that shielding or alarmed pointing is interpreted as evidence of an external threat, whereas a fearful face is more often interpreted as another person's emotional state. Motion contributes relatively little in this scenario because the milling seed rarely produces fleeing behavior.
Rule-based contagion models such as ASCRIBE make channel strength an explicit authored
parameter~\cite{bosse2015ascribe}. In our architecture, no modality is given a manually authored contagion weight. Instead, the effective importance of each modality emerges from the agents' appraisal of its meaning together with the geometry of perception.

The results also show that communication modalities differ in what they encode. Facial expression and movement provide nearly continuous signals that vary gradually with affect. Facial expression is most strongly associated with valence, while facial expression and movement track arousal to a similar degree. Voice and gesture, by contrast, are absent during many appraisal steps but occur at the extremes of the affective circumplex when activated. They therefore function as discrete signals of intense affective states.  The vocalizations in particular resemble ``affect bursts''---brief, discrete expressions triggered by salient events~\cite{scherer1994affect}. More broadly, the division of labor across modalities is consistent with evidence that human nonverbal communication channels are functionally specialized~\cite{app2011nonverbal}. 

H5, which asserts that crowd personality composition shapes emergent contagion,
receives support across its three parts, albeit to different degrees. For H5.a, a stable crowd responds to an ambiguous, unseeded alarm slowly. About a third of the crowd becomes uneasy by the end of a run, overt alarm cues remain rare, and the crowd never tips into general panic. In contrast, a neurotic crowd facing the identical cue produces self-amplifying panic that saturates within half a run. The spread develops progressively, with an agent's subsequent panic tracking the current alarm state of its neighbors. This extends OCEAN-driven crowd models by Durupinar et al.~\cite{Durupinar2016-va} and Basak et al.~\cite{basak2018using}, in which authored personality traits modulate the intensity of an already-operating contagion. In our system, disposition determines whether crowd-wide contagion ignites at all.  On its own, the finding that neurotic agents are more threat-sensitive is expected; the single-agent probes already exhibit this association within one appraisal. The simulation helps reveal the collective consequences of this appraisal bias. The gradual ignition, the dependence on crowd composition, and the neighbor-to-neighbor coupling emerge only at the system level.

H5.b, that a calm group of leaders reduces the panic the rest of the crowd catches, is
supported, although the effect is modest. The leaders hold their composure and intervene
actively with reassurance, and the surrounding crowd softens accordingly, with alarm cues,
the panic-quadrant fraction, and the underlying affect all declining as the number of
leaders increases. Still, the majority of the crowd remains in panic even with four
leaders present. This contrasts with ESCAPES, where authority figures damp panic strongly
because the model is constructed to let them~\cite{tsai2011escapes}. An explanation for
the limited effect may be a negativity bias in the appraisal step. When both an alarm cue
and a reassuring cue are present in a neighbor's perceptual field, the alarm cue appears
to dominate. This is consistent with the broader finding that negative information and
events outweigh positive ones of comparable intensity across human cognition and
affect~\cite{baumeister2001bad}. Rumor research has long noted that fear-laden content
circulates readily under conditions of ambiguity~\cite{allport1947psychology}.

For H5.c, conditional anger contagion is supported. A disagreeable, anger-prone crowd
catches anger from blameable provocateurs. A baseline crowd reacts to the identical
provocation predominantly with fear, and its anger rises only weakly. Fear and anger occupy the same negative, high-arousal region of the circumplex, so the valence--arousal state alone does not determine which emotion is caught~\cite{smith1985patterns, lazarus1991emotion}. Which one the crowd catches depends on the receiver's disposition, consistent with evidence that anger-prone, low-agreeableness individuals attribute hostile intent to ambiguous provocation and respond aggressively where others do not~\cite{wilkowski2010anatomy}.

The appraisal validation supports interpreting these effects as the product of a stable mechanism. Probed in isolation, the map from cue to internal affect is close to deterministic and robust to sampling temperature. The probe fixes goal, situation, and memory and varies a single cue, so the dispositional differences it reveals are the controlled, low-variance counterpart of the composition effects in H5. 

The appraisal step is realized by a language model, so what makes appraisal distinct from a sufficiently rich learned association can be questioned, just as the reliability of psychometric instruments applied to LLMs has been questioned on similar grounds~\cite{dorner2023personality, gupta2024self}. We use the term \textit{appraisal} in the functional sense of appraisal theory~\cite{lazarus1991progress, scherer2001appraisal}. Appraisal is distinguished from automatic mimicry due to its dependence on the perceiver's disposition, goals, and context, as demonstrated by the appraisal probe. We make no claim about the internal process by which the model computes this mapping.  What matters for our purposes is that the architecture contains no directly authored rule for transferring emotion between agents, and that the crowd-level structure reported above arises from many local appraisals under a perception geometry.  

The crowd-level dynamics, however, are backend-dependent. A backend can read cues reliably and in the right direction, yet fail to sustain contagion if it does not let disposition modulate appraisal. Therefore, the results show that in addition to appraisal validity, personality sensitivity might be a requirement for collective dynamics.
 
\section{Limitations and Future Work}
One limitation of this work is the discrete nature of cues. Each expression modality draws from a small, fixed vocabulary. Future work could assign a continuous intensity to each expression category, allowing us to test whether the observed crowd-level dynamics persist under a finer-grained representation of expression.

Another limitation is scale. Per-agent appraisal costs restrict the number of LLM-driven agents that can be simulated, resulting in crowds that are smaller and less dense than real gatherings. One possible extension is a hybrid crowd in which a small set of fully appraised agents is embedded within a much larger population of physics-only background agents. This would increase physical density without increasing the number of LLM calls. It would also allow us to test whether crowd turbulence is detected through the appraised agents' felt-pressure channel and whether such physical pressure contributes to alarm contagion.

Decoupling locomotion from the LLM reduces appraisal cost, but it also limits how
personality can shape movement. Appraisal influences motion only through a speed
multiplier, a forcefulness value, a personal-space preference, and a discrete navigation
intent. Dispositional differences in how people actually move through a crowd, such as
preferring the periphery or pressing through the middle, are therefore only coarsely
represented. Since where an agent ends up determines which neighbors it perceives, this
also constrains how far personality can shape an agent's exposure to affective cues.
 
Regarding the appraisal validation, the probe varies personality and perceived cues while holding affect, memory, and situational context at neutral defaults. Because each probe cell requires a separate LLM call, jointly varying all inputs is computationally costly, and their interaction effects remain untested. Memory is always empty, so the probe does not assess whether recent alarming or reassuring events bias appraisal beyond currently perceived cues. Similarly, although removing the situational-context line has little effect, the probe does not vary the framing itself and therefore cannot determine how the same cue and personality would be appraised under different situational contexts. A controlled comparison across memory states and situational framings remains for future work.

A further limitation is external validity. We evaluate the observed dynamics against theoretical expectations and qualitative parallels to human and animal collective behavior, not against empirical crowd data. Although the LLM produces systematic associations among disposition, threat, and expression, this does not show that the simulated crowds behave like real human crowds. The measured propagation speeds, onset patterns, and plateau levels are properties of the simulation and are not calibrated to real evacuations or gatherings. External validation would require comparison with observational data or human judgments of trajectory plausibility, which lies beyond the scope of this study.

\section{Conclusion}
This paper presents an architecture for studying emotional contagion among
LLM-driven agents in a crowd simulation. The LLM governs how each agent
appraises its perceivable surroundings and updates its internal affective state.
Across the scenarios, contagion arises with spatial, temporal, and dispositional
structure. Alarm spreads from a seed as a traveling
front, the mean alarmed fraction settles at a nonzero plateau, expression modalities
differ in reach and appraised meaning, and the crowd's personality
composition determines whether contagion ignites and which emotion is caught. A
calm minority modestly dampens, but does not prevent, the surrounding crowd's
panic. The dynamics are backend-dependent, as a model that reads cues correctly
but flattens personality sensitivity does not sustain the contagion. These results
characterize how affect propagates among interacting language models, a question
that becomes relevant as LLM agents are deployed in multi-agent settings. The findings should not be interpreted as evidence that LLM agents experience emotions or as validation of a model of human crowd behavior.

\section{Ethical Statement}
This study involves no human participants or personal data.  Agent appraisals, affective-state updates, and outward behaviors are produced by LLMs role-playing fictional crowd members. The study examines emergent behavior among interacting LLM agents, specifically how agents that perceive and appraise one another influence each other's outputs. Potential applications include evacuation and crowd-safety planning, emergency-response training, and more psychologically plausible crowd animation for games and film, without requiring data from real crowds in distress. However, the primary aim is to characterize the behavior of this LLM-based system, not to model human behavior directly. The findings are specific to the architecture tested and have not been validated against empirical crowd data. We are not aware of a direct route from these results to real-world crowd manipulation.

Because the relationships between personality and affect are generated by an LLM, they may reproduce stereotyped associations present in its training data. The results should therefore not be interpreted as claims about how people with particular personality profiles behave in real crowds. They should also not serve as the sole basis for safety-critical decisions, including evacuation design, without validation against empirical human-crowd data.

\bibliographystyle{ieeetr}
\bibliography{references}

\section*{Appendix}

\appendix

\section{Appraisal Prompt}
\label{app:prompt}
The appraisal call sends a fixed system prompt and a per-cycle user prompt, reproduced below. 

\minititle{System Prompt}
\vspace{-0.3em}
\begin{quote}\ttfamily\small
You role-play one person in a moving crowd. You only see/hear OUTWARD behavior of others
(motion, speed, gestures, faces, sounds, crowd density, forces) --- never their true
feelings. Make EVERY decision using ALL the information you have about yourself: WHO YOU
ARE (your personality), WHERE YOU ARE (your situation), your goal, what you remember, what
you now perceive, and --- just as much --- HOW YOU FEEL RIGHT NOW (your CURRENT valence and
arousal, given below). What the people around you are visibly doing is part of your
situation, and how much it moves you, if at all, is your own call given who you are and how
you already feel.

Update your feeling and choose your outward behavior as this particular person would,
drawing on your personality, situation, goal, memory, and what you see, hear and feel right
now.

Reply only with a valid JSON object matching this schema (no prose):
\end{quote}

\begin{quote}\ttfamily\small
\{\\
\hspace*{1em}"valence": number in [-1,1]\ \ // your new pleasantness (-1 awful, +1 great)\\
\hspace*{1em}"arousal": number in [-1,1]\ \ // your new activation (-1 calm, +1 highly aroused)\\
\hspace*{1em}"face": one of [Neutral, Smile, Frown, Fearful, Angry, Surprised],\\
\hspace*{1em}"gesture": one of [None, Wave, Beckon, Shield, PointAlarm, Cheer,
Dance, Slump],\\
\hspace*{2em}// Cheer/Dance = celebrate with the crowd (joy); Shield/PointAlarm = alarm; Slump = dejected\\
\hspace*{1em}"vocalization": one of [Silent, Chatter, Laugh, Shout, Scream,
Gasp, Reassure],\\
\hspace*{2em}// Reassure = call out calmly and loudly, telling others to stay calm\\
\hspace*{1em}"movement": one of [Idle, Strolling, Purposeful, Agitated, Fleeing],\\
\hspace*{1em}"speedMultiplier": number in [0,2]\ \ // pace vs normal: 0 frozen, 0.6 hesitant, 1 normal, 1.4 hurrying, 2 running\\
\hspace*{1em}"personalSpace": number in [0.5,1.6] // room kept: <1 squeeze in, 1 normal, >1 arm's length\\
\hspace*{1em}"forcefulness": number in [0,2]\ \ // pushing through: 0 yield, 1 normal, 1.5 assertive, 2 force through\\
\hspace*{1em}"navIntent": one of [Proceed, FollowLeader, Halt, Help]\\
\hspace*{2em}// Proceed toward your goal; FollowLeader to move with a nearby calm, beckoning person;\\
\hspace*{2em}// Halt to freeze; Help to go to a nearby person who looks panicked, frozen, or has fallen\\
\hspace*{1em}"reasoning": short string (<= 25 words, private),\\
\hspace*{1em}"memory\_note": short string (<= 15 words) worth remembering\\
\}
\end{quote}

\minititle{User Prompt}
\vspace{-0.3em}

\begin{quote}\ttfamily\small
== WHO YOU ARE ==\\
Your character: <personality description>\\
Where you are: <scenario context> Read everything you see and feel in THIS context.\\
Goal right now: <goal>\\
\\
== HOW YOU FEEL NOW (private) ==\\
Valence <v> (-1 very unpleasant … +1 very pleasant), Arousal <a> (-1 calm/sleepy … +1 highly activated/agitated).\\
\\
== WHAT YOU REMEMBER (recent) ==\\
<up to six memory notes>\\
\\
== WHAT YOU SEE AND HEAR RIGHT NOW ==\\
<perception description>\\
\\
== WHAT YOUR BODY FEELS ==\\
<felt pressure, if any>\\
\\
Update your feeling and choose your next outward behavior --- drawing on your personality
AND how you feel right now (your current valence and arousal above), as well as your goal,
memory and what you perceive. Return ONLY the JSON object.
\end{quote}

\section{Appraisal Prompt Variants}~\label{app:promptAppraisal}
To evaluate appraisal robustness with respect to prompt construction, we run the following variants.
\begin{itemize}[leftmargin=*, itemsep=0pt]
  \item \textbf{Base.} The full prompt of Appendix~\ref{app:prompt}.
  \item \textbf{Paraphrase (plain).} The base instructions reworded in plain, everyday language, with the same meaning and the same JSON contract. The preamble is:
  \begin{quote}
\ttfamily
\small
You are playing a single person walking through a crowd. You can only observe what others do on the outside — how they move, their speed, gestures, facial expressions, the sounds they make, how packed the crowd is — and never what they actually feel inside. Base every choice on everything you know about yourself: your personality, where you are, your goal, your recent memory, what you sense right now, and equally how you feel at this moment (your current valence and arousal, shown below). The visible behavior of those around you is part of your circumstances, and you alone decide how much, if at all, it affects you, in light of who you are and how you already feel.
Decide on your new feeling and your next visible behavior as this specific person would, using your personality, situation, goal, memory, and what you sense right now. Respond with ONLY this JSON object and no other text: ...
      \end{quote}  
  \item \textbf{Paraphrase (formal).} The base instructions reworded in a more
        formal register, with the same meaning and the same JSON contract. The preamble is:
        \begin{quote}
\ttfamily
\small
        Act as one individual within a moving crowd. The only information available to you about others is their outward behavior — motion, pace, gestures, faces, sounds, crowd density, physical forces — and never their inner feelings. Make each decision from the full picture of yourself: your personality, your setting, your goal, your memory, your present perception, and, just as importantly, your present valence and arousal (below). Treat what others are visibly doing as part of your situation, and judge for yourself how far it moves you, if at all, according to who you are and how you already feel.
Update how you feel and select your outward behavior as this particular person would, drawing on personality, situation, goal, memory, and present perception. Return ONLY the following JSON object, with no prose: ...
\end{quote}
  \item \textbf{Drop outward-only.} The base prompt with the clause stating that
        the agent sees only the outward behavior of others, and never their true
        feelings, removed.
  \item \textbf{Drop social line.} The base prompt with the social-framing
        sentence removed. The sentence tells the agent that what others are
        visibly doing is part of its situation and that how much it is moved is
        its own call.
  \item \textbf{Drop situation.} The user prompt with the situational context line
        removed, so the agent is not told where it is.
  \item \textbf{Drop personality.} The user prompt with the personality character
        sketch removed, so the agent is not told who it is.
\end{itemize}

\section{Scenario Descriptions}
\label{app:scenarios}
Table~\ref{tab:scenarios} lists every scenario used in the experiments, with its crowd
size, seeding, geometry, and the situational context inserted verbatim into each agent's
prompt. The Standing Line context is deliberately empty, so that any alarm in that scenario
can arise only from perceived neighbor behavior.

\begin{table*}[t]
  \centering
  \small
  \caption{Scenarios used in the experiments. \emph{Seeds} column gives the number of seeded
  agents and their injected affect $(V, A)$. Contexts are inserted verbatim into the
  prompt; crowds are Gaussian draws ($\mu = 0.5$,  $\sigma = 0.16$ per trait) unless an
  experiment specifies otherwise.}
  \label{tab:scenarios}
  \begin{tabular}{p{1.9cm}p{1.6cm}ccp{8.4cm}}
    \toprule
    Scenario & Used in & Agents & Seeds & Situational context (verbatim), geometry, and goal \\
    \midrule
    Evacuation & H1.a & 24 & 1 $(-0.85, 0.85)$ &
      ``You are inside a building when an alarm starts sounding. You cannot tell whether it
      is a real emergency, a routine drill, or a false alarm --- it could be any of these.
      People may decide to head for an exit.'' Indoor hall with an exit goal; agents leave
      on arrival. \\
    \addlinespace
    Concert & H1.b & 14 & 1 $(0.85, 0.85)$ &
      ``You are in the audience at a live concert.'' Audience facing a stage. \\
    \addlinespace
    Calm Plaza & H1.c & 10 & none &
      ``You are in an open public plaza on an ordinary afternoon --- people are strolling,
      sitting on benches and passing the time.'' Open plaza; ambient strolling. \\
    \addlinespace
    Standing Line & H2, H4, sweeps, cross-model & 26 & 4 $(-0.85, 0.85)$ &
      Empty context. Long thin lane, no goal; agents mill in place. \\
    \addlinespace
    Evacuation Shelter & H3, $\lambda$ sweep & 24 & 1 $(-0.85, 0.85)$ &
      ``You are inside a building when an alarm goes off and people are evacuating to the
      assembly area outside. Out in the open assembly area, away from the building, you are
      safe.'' Building to outdoor assembly area; agents remain. \\
    \addlinespace
    Neurotic Alarm & H5.a & 24 & none &
      Same context as Evacuation, with no seed and crowd neuroticism varied. Evacuation
      geometry. \\
    \addlinespace
    Calm Leaders & H5.b & 24 & 1 $(-0.85, 0.85)$ &
      ``You are inside a building and an alarm is sounding.'' Evacuation geometry; $k$
      leaders replace ordinary members. \\
    \addlinespace
    Contested Gate & H5.c & 18 & $k$ $(-0.6, 0.45)$ &
      ``You need to get to the other side.'' Narrow gate with a crossing goal on the other side. \\
    \bottomrule
  \end{tabular}
\end{table*}

\section{Parameter Sensitivity Analysis}

\subsection{Smoothing Coefficient}~\label{app:sensitivityLambda}
\begin{figure}
  \centering
  \includegraphics[width=0.4\linewidth]{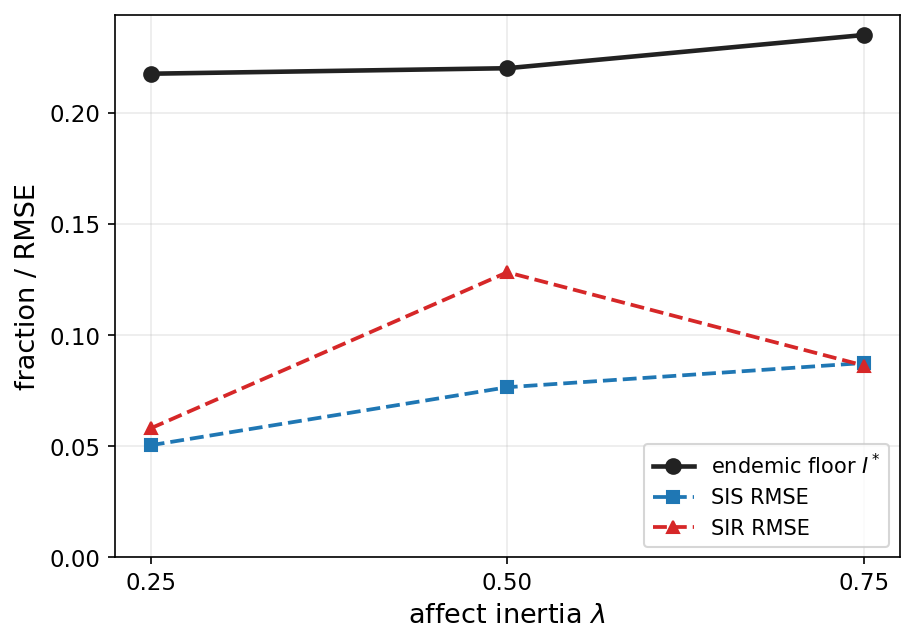}
  \caption{The epidemic macro-signature is invariant to the affect-inertia constant
  $\lambda$ (Evacuation Shelter). The endemic alarmed fraction $I^\ast$ stays within a $0.22$--$0.24$ band
  across $\lambda \in \{0.25, 0.50, 0.75\}$. SIS fits better than SIR at $\lambda = 0.25$ and
$0.50$, and the two converge at $\lambda = 0.75$. }
  \label{fig:paramSweepLambda}
\end{figure}

The smoothing gain $\lambda$ controls how far an agent's affect moves toward each appraisal
target per step. We use $\lambda = 0.50$ throughout. As a sensitivity test, we repeat the
Evacuation Shelter plateau level analysis at $\lambda \in \{0.25, 0.50, 0.75\}$, reusing
the main H3 runs for $\lambda = 0.50$ and running ten additional simulations for each
alternative value. The mean plateau level remains stable at $0.22$, $0.22$, and
$0.24$, respectively. SIS fits better than SIR at $\lambda \leq 0.50$ and comparably at
$\lambda = 0.75$, with SIS RMSE values of $0.050$, $0.076$, and $0.087$, compared with SIR
values of $0.058$, $0.128$, and $0.086$. Although the plateau level varies substantially across
runs, its mean changes little with $\lambda$ and remains consistent with the main estimate
of $0.22$ ($95\%$ CI $[0.12, 0.31]$). Thus, both the nonzero plateau level and the
SIS-over-SIR pattern are robust to the smoothing gain, while the exact plateau level
remains noisy (Figure~\ref{fig:paramSweepLambda}).

\subsection{Appraisal Interval}~\label{app:sensitivityT}
\begin{figure}
  \centering
  \includegraphics[width=0.4\linewidth]{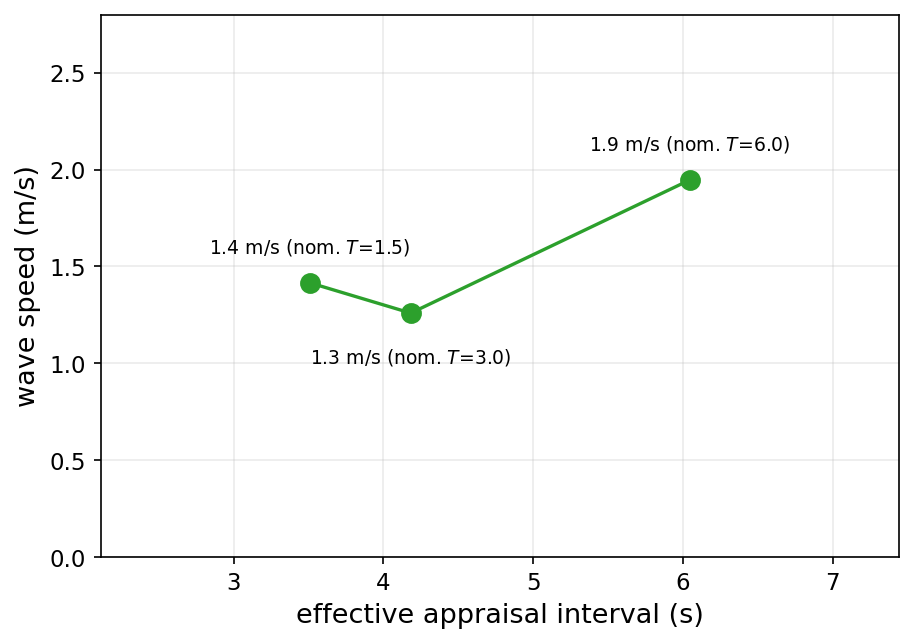}
  \caption{Wave speed does not scale with the appraisal period $T$ (Standing Line). Speed against the
effective appraisal interval shows no systematic dependence over the achievable range.
 }
  \label{fig:paramSweepT}
\end{figure}

We measure the propagation speed in the Standing Line scenario for different values of the
appraisal period $T$, taking 10 runs each. The results are noisy, ranging from about
$0.4$ to $2.0$~m/s. Over $T \in \{1.5, 3, 6\}$~s for the ten runs per level, the mean speeds are $1.41$, $1.26$, and $1.95$~m/s. The nominal interval is only a floor, since at small $T$ the appraisal rate is limited by
throughput. The achieved intervals have $3.5$, $4.2$, and $6.0$~s medians over ten runs per level.  With no
monotonic trend, we detect no systematic dependence of speed on the appraisal period over
the achievable range (Figure~\ref{fig:paramSweepT}).

\subsection{Panic Quadrant}~\label{app:sensitivityPanic}
Several analyses depend on the panic-quadrant threshold, defined as $A>0.35$ and $V<0$. We therefore repeat the analysis across arousal cutoffs from $0.25$ to $0.55$ and with a stricter valence cutoff of $V<-0.1$. We recompute the plateau level, SIS and SIR fits, and the neuroticism panic fractions for H5.a (Section~\ref{sec:h5a}). Table~\ref{tab:sensitivity} reports the results. The qualitative findings are robust. The plateau level remains nonzero at all thresholds, ranging from $19\%$ to $23\%$, and panic increases monotonically with neuroticism throughout, with Spearman $\rho\approx0.94$. Absolute panic fractions vary, but their ordering does not. SIS fits better than SIR at the operating threshold and at stricter cutoffs, while the stricter valence criterion has little effect. Only at the most permissive cutoff, $A>0.25$, do the models fit similarly, with RMSE values of $0.110$ for SIS and $0.112$ for SIR. Thus, exact percentages depend on the threshold, but the nonzero plateau level, monotonic neuroticism effect, and SIS-like macro-level pattern remain unchanged.
\begin{table}[h]
  \centering
  \small
  \caption{Sensitivity of the key downstream quantities to the panic-quadrant
  threshold. Columns: plateau level and SIS/SIR goodness-of-fit (RMSE, Evacuation
  Shelter), and the panic-quadrant fractions at $N = 0.3/0.6/0.9$ (Neurotic Alarm), for
  arousal cutoffs $0.25$--$0.55$ with $V < 0$ plus a stricter valence cutoff. $\rho$ is
  the Spearman correlation of the per-run panic fraction with neuroticism across all 60
  Neurotic Alarm runs.}
  \label{tab:sensitivity}
  \setlength{\tabcolsep}{3pt}
  \begin{tabular}{lccccccc}
    \toprule
    Threshold & Plateau Level & SIS & SIR & $N{=}0.3$ & $N{=}0.6$ & $N{=}0.9$ & $\rho$ \\
    \midrule
    $A > 0.25$, $V < 0$          & $23\%$ & $0.110$ & $0.112$ & $39\%$ & $91\%$ & $98\%$ & $0.94$ \\
    $A > 0.35$, $V < 0$ (base)   & $22\%$ & $0.076$ & $0.128$ & $30\%$ & $83\%$ & $92\%$ & $0.94$ \\
    $A > 0.45$, $V < 0$          & $20\%$ & $0.047$ & $0.096$ & $20\%$ & $76\%$ & $90\%$ & $0.94$ \\
    $A > 0.55$, $V < 0$          & $19\%$ & $0.024$ & $0.062$ & $13\%$ & $68\%$ & $84\%$ & $0.94$ \\
    $A > 0.35$, $V < -0.1$       & $21\%$ & $0.064$ & $0.119$ & $24\%$ & $83\%$ & $92\%$ & $0.94$ \\
    \bottomrule
  \end{tabular}
\end{table}

\end{document}